\documentclass[sigconf,nonacm,dvipsnames]{acmart}
\usepackage{svg}
\usepackage{subcaption}
\usepackage{listings}
\usepackage{forest}
\usepackage{tikz, tikz-qtree}
\usepackage{spverbatim}
\usepackage{multirow}
\usepackage{stfloats}

\lstdefinelanguage{DSL}{
  morekeywords=[1]{IF, PROGN, WHILE},
  keywordstyle=[1]\color{black}\bfseries,
  morekeywords=[2]{AND, OR, NOT, BOOL},
  keywordstyle=[2]\color{blue},
  morekeywords=[3]{IS_WALL, IS_RESOURCE, IS_EMPTY, IS_HAZARD, PRED},
  keywordstyle=[3]\color{ForestGreen},
  morekeywords=[4]{FORWARD, LEFT, RIGHT, DIR},
  keywordstyle=[4]\color{orange},
  morekeywords=[5]{TURN_LEFT, TURN_RIGHT, MOVE_FORWARD, NO_OP, ACTION},
  keywordstyle=[5]\color{red},
  identifierstyle=\color{black},
  sensitive=false,                              
  comment=[l]{//},                              
  morecomment=[s]{/*}{*/},                      
  commentstyle=\color{purple}\ttfamily,
  stringstyle=\color{red}\ttfamily,
  morestring=[b]',
  morestring=[b]"
}

\AtBeginDocument{%
  }

\begin{document}

\title{Mutation Without Variation: Convergence Dynamics in LLM-Driven Program Evolution}

\author{Can Gurkan}
\authornote{Corresponding Author}
\email{gurkan@u.northwestern.edu}
\orcid{0009-0001-3623-6715}
\affiliation{%
  \institution{Northwestern University}
  \city{Evanston}
  \state{IL}
  \country{USA}
}

\author{Forrest Stonedahl}
\affiliation{%
  \institution{Augustana College}
  \city{Rock Island}
  \state{IL}
  \country{USA}
}

\author{Uri Wilensky}
\affiliation{%
  \institution{Northwestern University}
  \city{Evanston}
  \state{IL}
  \country{USA}
}

\renewcommand{\shortauthors}{Gurkan et al.}

\begin{abstract}
  When an LLM repeatedly mutates a program, does it explore new forms or circle back to the same ones? We study this question by analyzing LLM-driven mutation chains in the absence of selection pressure within a domain-specific language, varying prompt design, model family, and stochastic replication. We find that LLM-based mutation consistently converges toward restricted attractor regions in program space. Convergence is especially severe at the structural level: in 87\% of chains, over 93\% of mutations revisit a previously seen structural form, with most variation confined to terminal substitutions within recurring templates. Cycle analysis reveals short cycles and self-loops dominating the transition structure. The rate of convergence varies with prompt wording and model choice, but the phenomenon is robust across conditions. A classical GP subtree mutation operator does not exhibit comparable convergence, suggesting that the effect is intrinsic to the LLM mutation pipeline. These findings reveal a tension at the heart of LLM-driven program evolution: the same capabilities that enable semantics-aware program transformation also carry a systematic bias toward structural homogeneity that must be accounted for if such systems are to sustain open-ended exploration. Source code is available at \url{https://github.com/can-gurkan/lmca}.
\end{abstract}



\maketitle

\footnotetext{Accepted to the Genetic and Evolutionary Computation Conference (GECCO '26) Workshop on Large Language Models for and with Evolutionary Computation.}

\section{Introduction}

\begin{figure}
    \centering
    \includegraphics[width=\linewidth]{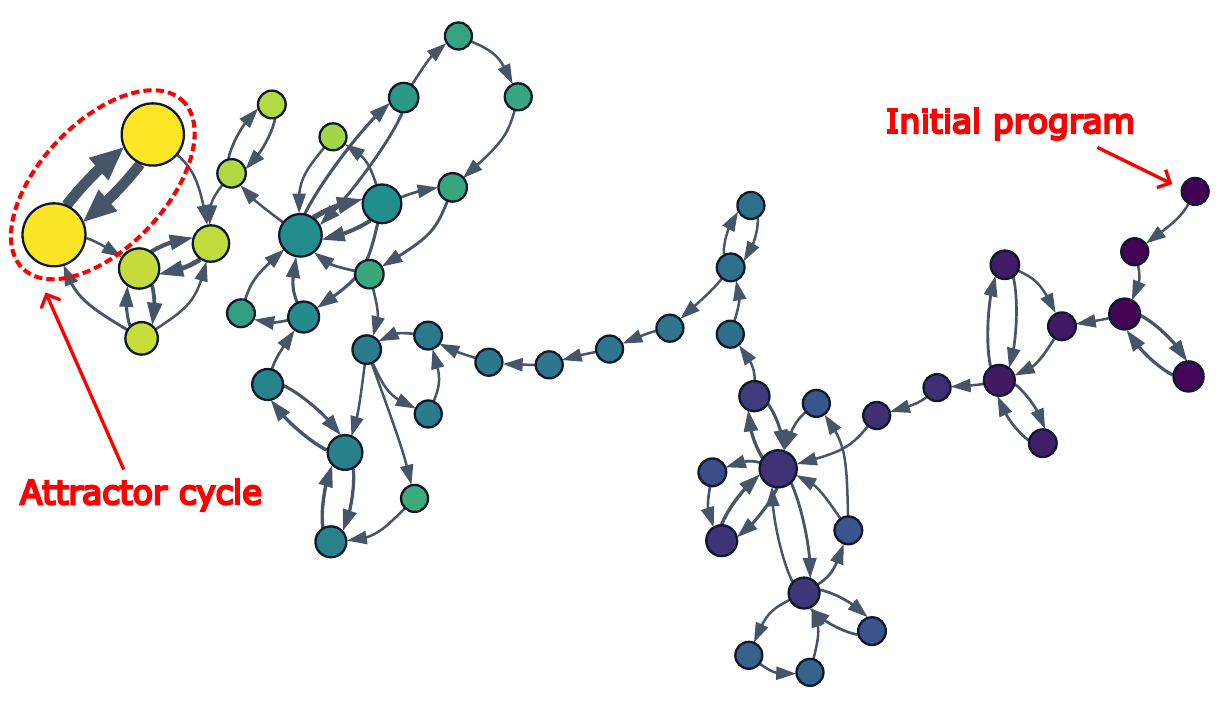}
    \caption{Example transition graph from a single LLM mutation chain. Nodes represent unique program states, with size proportional to visit frequency and color indicating discovery order. The trajectory passes through a transient sequence before settling into a 2-cycle attractor.}
    \Description{A directed graph showing program states as nodes connected by edges representing mutations. Several nodes form a linear chain from the initial program, leading to two large nodes that alternate in a 2-cycle. Node sizes vary, with the two attractor nodes being the largest.}
    \label{fig:sample-mutation-trajectory}
\end{figure}

In the conclusion of \textit{On the Origin of Species}~\cite{darwin1859}, Charles Darwin wrote, ``...from so simple a beginning endless forms most beautiful and most wonderful have been, and are being, evolved’’. This vision of open-ended innovation has long served as both inspiration and aspiration for evolutionary computation. Evolutionary algorithms (EAs) rely on the premise that simple variation operators, applied iteratively, can give rise to a vast and diverse space of candidate solutions. Among these operators, mutation plays a central role, continually injecting new genetic material and enabling populations to explore beyond their current configurations. Both the rate at which variation is introduced and the magnitude of individual changes shape the trajectory of the search, and maintaining this balance has always been of central importance to a successful EA~\cite{EA-Banzhaf}.

Recently, advances in large language models (LLMs) have enabled a new paradigm of LLM-driven discovery, in which models iteratively generate and refine diverse candidate solutions~\cite{LLM_recent_advances}. Ushered in by the seminal work of Lehman et al.~\cite{ELM}, this line of research embeds LLMs within iterative optimization loops, where they generate, modify, and recombine solutions across successive generations~\cite{LLM_EA_review, LLM_survey_code, LLM_ES, LLM_BBO}. Such systems have shown promise in domains ranging from program synthesis to scientific discovery, at times producing solutions that appear to extend beyond the scope of their training distributions~\cite{LLM_funsearch, alphaevolve, shinkaevolve, LLM_ADAS, LLM_aiscientist, llm_promptbreeder, LLM_eureka, LLM_discopop, LLM_EOH, LLM_hyperparam_EA, partevo, ReEvo, llamea}.

A particularly active application of this paradigm is in genetic programming (GP)~\cite{koza:book}, where LLMs are employed to generate syntactically valid code, enhancing the capabilities of evolutionary operators~\cite{ELM,LMX,LLM_OpenELM,LLM_GP_code_Hemberg2024, LEAR}. This integration has given rise to LLM-driven genetic programming (LLM-GP), in which generative models serve as variation operators within the evolutionary loop~\cite{LLM_GP_review}.

A central component of the LLM-GP framework is the use of LLMs as mutation operators for programs. Unlike traditional syntactic perturbations, LLM-based mutations are semantics-aware and capable of producing coherent, structured transformations that are often directly applicable to modern programming languages~\cite{ELM, LEAR, LLM_GP_code_Hemberg2024}. These properties suggest that LLMs provide a qualitatively different mechanism for traversing program space, one that is more guided and potentially more effective than conventional mutation operators.

However, this promise raises a more fundamental question that remains largely unexamined. When an LLM is used repeatedly as a mutation operator, how does it traverse the space of programs? Does it continue to generate novel forms in the spirit of Darwin’s observation, or does it instead exhibit a tendency to converge toward particular regions of the space, gradually limiting the diversity of representations? If LLM-driven mutation introduces implicit biases that steer programs toward canonical structures favored by the model’s training distribution, such biases could shape evolutionary search trajectories in ways that are not immediately apparent, potentially restricting exploration even in the absence of explicit constraints.

Existing work has largely examined LLMs within optimization loops, where their effectiveness is measured in terms of solution quality~\cite{LLM_EA_review, vanstein2025behaviourspace, vanstein2025codeevolgraphs, LLM_survey_code, LLM_ES, LLM_BBO, LLM_funsearch, alphaevolve, shinkaevolve, LLM_ADAS, LLM_aiscientist, llm_promptbreeder, LLM_eureka, LLM_discopop, LLM_EOH, LLM_hyperparam_EA, partevo, ReEvo, llamea, ELM,LMX,LLM_OpenELM,LLM_GP_code_Hemberg2024, LEAR, LLM_GP_review, digitalredqueen}. This perspective obscures the intrinsic behavior of the mutation operator itself: the presence of selection pressure makes it difficult to disentangle whether observed convergence arises from the fitness landscape or from biases inherent to the LLM. To understand the role of the mutation operator in isolation, it is therefore necessary to study its behavior independently of selection, focusing on how it transforms programs over repeated application.

The question of how repeated application shapes LLM outputs connects to a broader class of iterative LLM systems, including chain-of-thought reasoning~\cite{llm_chain_of_thought22}, self-refinement pipelines~\cite{llm_selfrefine23}, and autonomous agents~\cite{LLM_ADAS, wang2024voyager, yao2023react}, in which model outputs are recursively fed back as inputs. Prior work has identified drift toward high-probability outputs and loss of diversity in text~\cite{perez2024telephone, wang2025attractorcycles, tacheny2025agentic, tripto2024theseus, mohamed2025brokentelephone} and image~\cite{hintze2025imagemotifs, mollah2025visualtelephone} domains, but analogous analyses in program space remain limited, despite code providing a structured domain in which convergence dynamics can be rigorously measured.

In this work, we study LLM-driven mutation as a stochastic, semantics-aware rewriting process over program space. We analyze mutation chains in the absence of selection pressure in order to isolate the intrinsic dynamics of the operator. Our central objective is to determine whether repeated application of LLM-based mutation promotes structural diversity through the generation of novel genetic material, or instead induces convergence toward restricted regions of the program space, potentially exhibiting attractor-like behavior. To enable precise structural analysis, we conduct these experiments within a constrained, strongly typed domain-specific language (DSL) that bounds the space of valid programs. Our investigation is purely genotypic: we analyze the structural properties of programs independent of their behavior, ensuring that the dynamics we observe reflect the operator's intrinsic biases rather than any property of the task environment or fitness landscape. By characterizing these dynamics across multiple models, prompts, and initial conditions, we aim to provide a quantitative account of how LLMs shape the evolution of program representations.

\subsection{Research Questions \& Contributions}
\label{sec:rqs}

Guided by these objectives, we address the following research questions:

\begin{enumerate}
\item Do LLM-driven mutation chains, in the absence of selection pressure, converge toward attractor regions in program space?
\item To what extent do these mutation processes reduce the diversity of program representations, and do patterns of convergence differ between high-level structural form and surface-level variation?
\item How sensitive are these dynamics to factors such as prompt design, initial program structure, and model family?
\end{enumerate}

To answer these questions, we make the following contributions:

\begin{itemize}
\item We introduce a framework for studying LLM-driven mutation as a dynamical system over program space, enabling the analysis of mutation chains independent of optimization objectives.
\item We provide empirical evidence of convergence and diversity stagnation in neutral LLM-driven mutation chains, identifying recurring patterns of change at both the structural and surface level, as well as oscillatory behaviors, across multiple experimental conditions.
\item We conduct a systematic analysis of how these dynamics vary across models, prompts, and initial conditions within a constrained DSL setting, revealing the extent to which convergence is a robust property of the LLM mutation pipeline rather than an artifact of any single configuration.
\end{itemize}

\section{Related Work}

We situate this work at the intersection of three lines of research: the integration of LLMs into evolutionary computation, the study of convergence dynamics in iterative LLM systems, and the role of drift in program evolution.

\subsection{LLMs in Evolutionary Computation}

The emergence of large language models as components of evolutionary search has reshaped evolutionary computation, enabling a new class of systems in which generative models augment or replace traditional variation operators~\cite{LLM_GP_review, LLM_EA_review}. Lehman et al.~\cite{ELM} demonstrated that code-generating LLMs can serve as effective mutation operators for genetic programming by approximating semantics-aware program transformations. This insight has since produced systems for mathematical program discovery~\cite{LLM_funsearch}, large-scale algorithmic optimization~\cite{alphaevolve, shinkaevolve}, heuristic design~\cite{LLM_EOH, ReEvo, llamea}, control policy synthesis~\cite{guo2026codeevol}, and GP-based code evolution~\cite{LLM_GP_code_Hemberg2024, LEAR}. Code Evolution Graphs~\cite{vanstein2025codeevolgraphs, vanstein2025behaviourspace} have further provided tools for analyzing how these systems traverse algorithm space.

Despite the successes of LLM-driven evolutionary systems, a recurring observation is that LLM-driven mutation operators appear to exert systematic pressure toward particular program structures. Analyses of the LLaMEA framework’s behavior space~\cite{vanstein2025behaviourspace} find that independent runs tend to cluster toward structurally similar solutions. Digital Red Queen~\cite{digitalredqueen} similarly observes convergence toward generalist strategies across adversarially evolving populations despite diverse initializations. Indeed, diversity-preserving mechanisms have accompanied LLM-driven mutation from the outset: the ELM framework of Lehman et al.~\cite{ELM} itself employed a quality-diversity algorithm, and subsequent systems have similarly incorporated diversity-preserving archives~\cite{digitalredqueen}, novelty-based rejection sampling~\cite{shinkaevolve}, and niching~\cite{partevo}. While these mechanisms treat convergence as a practical failure mode to be corrected, the intrinsic dynamics of the mutation operator in the absence of such mechanisms remain unexplored. This leaves open a fundamental question: what intrinsic dynamics does the LLM mutation operator induce in program space when decoupled from selection, and do these dynamics persist robustly across models and prompting strategies?

\subsection{Convergence Dynamics in Iterative LLM Systems}

Beyond evolutionary computation, convergence dynamics have been observed in iterative LLM systems across multiple domains and modalities. Perez et al.~\cite{perez2024telephone} study iterated LLM-to-LLM transmission chains modeled on telephone-game dynamics, showing that textual properties such as toxicity, positivity, and difficulty converge toward equilibrium values over successive rewrites. They introduce an attractor estimation method based on regression between initial and final property values, providing a quantitative notion of attractor position and strength. Mohamed et al.~\cite{mohamed2025brokentelephone} report similar cumulative distortion effects in iterative translation chains, where degradation depends on chain length and intermediate representations. Related studies on iterative LLM paraphrasing have documented similar cumulative transformation effects~\cite{tripto2024theseus, sadasivan2023recursive}. In the image domain, Hintze et al.~\cite{hintze2025imagemotifs} show that autonomous text-to-image-to-text loops converge to just 12 dominant visual motifs across 700 independent trajectories and seven temperature settings, providing striking evidence that convergence toward generic outputs is a robust property of iterative generative systems regardless of modality.

Building on these observations, several studies have formalized iterative LLM generation explicitly as a dynamical system. Wang et al.~\cite{wang2025attractorcycles} formalize successive paraphrasing as a discrete dynamical system and demonstrate that trajectories frequently converge not to fixed points but to 2-period limit cycles, introducing a 2-periodicity degree metric that captures alternating-cluster structure across iterations. Tacheny~\cite{tacheny2025agentic} analyzes agentic loops in which LLM outputs are recursively fed back as inputs, distinguishing between contractive dynamics that lead to attractors and divergent dynamics that promote exploration. Notably, the regime depends on prompt design, suggesting that convergence reflects a geometric property of the induced transformation that can be modulated through operator specification. Related theoretical work on model collapse~\cite{shumailov2023curse, shumailov2024nature} shows that contraction toward typicality and loss of distributional support can arise under recursive generation even at the training level, reinforcing the importance of understanding convergence dynamics at inference time as well. This dynamical systems perspective, treating iterative LLM generation as trajectories in a representational space with measurable contraction and periodicity, provides the conceptual framework adopted in the present work.

In the code domain specifically, Peitek et al.~\cite{peitek2026refactoring} provide one of the most direct empirical studies, analyzing multi-step code refactoring trajectories under repeated LLM application. Their results reveal a characteristic two-phase dynamic: an initial restructuring phase marked by substantial syntactic changes, followed by stabilization in which successive versions become increasingly similar. They also document prompt-dependent oscillatory behavior, in which code alternates between a small number of variants rather than converging to a fixed point. While this work targets code readability improvement rather than mutation in a genetic programming context, the observed dynamics offer compelling evidence that iterative LLM rewriting of code exhibits structured convergence patterns. However, a systematic investigation of convergence under LLM-driven mutation in program space, particularly without selection pressure and across variation in model and prompt design, has not yet been conducted.

\subsection{Drift in Program Evolution}

The concept of genetic drift has a long history in genetic programming, where neutral drift refers to sequences of mutations that preserve semantic behavior while allowing syntactic variation~\cite{GP_field_guide, OReilly1997GP-neutrality, miller2001-neutrality}. Atkinson et al.~\cite{atkinson2018snd} investigate this idea through equivalence-preserving transformations over program graphs, demonstrating that designed neutral moves can reshape evolutionary trajectories and alter the accessibility of regions in program space without invoking fitness-based selection. Their work utilizes a methodological principle we adopt here: by removing selection pressure, one can isolate and study the intrinsic properties of a mutation operator itself. The present work adopts this principle but shifts the setting: rather than rule-based, semantics-preserving rewrites, we apply LLM-driven mutations that are syntactically constrained but not restricted to preserve behavior. This allows us to study how trajectories move through program space under the operator’s bias, and to assess the extent to which genotypic diversity contracts in the absence of selection.

\subsection{This Work}

The present work sits at the intersection of these three lines of research. Prior studies of LLM-driven evolutionary systems have observed convergence but have not disentangled the contributions of selection pressure from the intrinsic limitations of the LLM-based mutation operator itself. Convergence dynamics have been formally studied in iterative text and image generation, but not in the context of LLM-driven mutation over program space. By isolating the genotypic-level dynamics of iterated LLM-driven mutation in the absence of selection pressure, and applying the dynamical systems framing developed for iterative LLM generation, this work provides a systematic empirical investigation of how LLM mutation operators shape trajectories through program space across multiple models and prompting strategies.


\section{Methods}

In this section, we describe the program representation, mutation operators, and trajectory analysis measures used throughout the experiments. The full source code is available at \url{https://github.com/can-gurkan/lmca}.

\subsection{Program Representation and DSL}

To facilitate direct comparison to classical tree-based GP, programs are expressed in a strongly typed, Lisp-like domain-specific language designed for gridworld agent control. The language comprises five categories of primitives, summarized in Table~\ref{tab:dsl}. Control flow operators support conditional branching, bounded looping, and sequential composition. Predicates are sensor queries, each parameterized by a relative direction.

\begin{table}[h]
\caption{DSL primitive categories.}
\begin{tabular}{ll}
\toprule
Category & Primitives \\
\midrule
Control flow & \texttt{IF}, \texttt{WHILE}, \texttt{PROGN} \\
Booleans & \textcolor{blue}{\texttt{AND}, \texttt{OR}, \texttt{NOT}} \\
Predicates & \textcolor{ForestGreen}{\texttt{IS\_WALL}, \texttt{IS\_EMPTY}, \texttt{IS\_HAZARD}, \texttt{IS\_RESOURCE} }\\
Directions & \textcolor{orange}{\texttt{FORWARD}, \texttt{LEFT}, \texttt{RIGHT}} \\
Actions & \textcolor{red}{\texttt{MOVE\_FORWARD}, \texttt{TURN\_LEFT}, \texttt{TURN\_RIGHT}, \texttt{NO\_OP}} \\
\bottomrule
\end{tabular}
\label{tab:dsl}
\end{table}

Every program is maintained as both a full program string, preserving all terminals and arguments, and a \emph{skeleton} representation that abstracts terminal-level details while preserving structural form. Specifically, action terminals are replaced with \texttt{ACTION}, direction terminals with \texttt{DIR}, boolean operators with \texttt{BOOL}, and predicates with \texttt{PRED}, while control-flow operators and tree structure are retained. Figure~\ref{fig:dsl_pair} depicts an example program and its skeleton; abstract syntax trees are shown in Appendix~\ref{app:ast}. This dual representation distinguishes structural changes from terminal-level substitutions.

\begin{figure*}
\centering
\begin{lstlisting}[language=DSL, basicstyle=\ttfamily\small, breaklines=true]
PROGN(IF(IS_WALL(FORWARD),TURN_LEFT,MOVE_FORWARD),WHILE(NOT(IS_RESOURCE(FORWARD)),TURN_RIGHT))
\end{lstlisting}
\begin{lstlisting}[language=DSL, basicstyle=\ttfamily\small, breaklines=true]
PROGN(IF( PRED  (  DIR  ),  ACTION,   ACTION  ),  WHILE(BOOL  (  PRED  (  DIR  ) ), ACTION ))
\end{lstlisting}
\caption{Example program in the DSL (top) and its corresponding skeleton representation (bottom). The skeleton abstracts terminal-level details while preserving control-flow structure.}
\Description{Two code listings side by side. The left shows a DSL program with specific primitives like IS\_WALL, TURN\_LEFT, and MOVE\_FORWARD. The right shows the same structure with terminals replaced by abstract categories: PRED, DIR, ACTION, and BOOL.}
\label{fig:dsl_pair}
\end{figure*}

\subsection{Mutation Operators}

\textbf{LLM Mutation.} Each mutation step prompts an LLM with the parent program and a specification of the allowed DSL primitives, instructing it to produce a single mutated variant. The full prompt templates are provided in Appendix~\ref{app:prompts}. After generation, the candidate is validated for syntactic well-formedness, primitive-set membership, and valid typing. If validation fails, the model is re-prompted with the failure reason and the invalid candidate for up to five retries. If all retries are exhausted, a fallback model is invoked with the same retry budget. Each chain step corresponds to one accepted, validated mutation. If both models exhaust their retry budgets the chain terminates, though this did not occur in any experiment reported here.

\noindent\textbf{Classical GP Mutation.} As a baseline, we run the same neutral chain procedure using a standard subtree mutation operator. At each step, a random node is selected in the program tree and replaced with a randomly generated subtree. A maximum tree depth of four is enforced to keep program sizes comparable to those produced by LLM mutation, preventing the count of unique programs from being inflated by unconstrained growth in program size.

\subsection{Trajectory Analysis Measures}

Our primary measure of convergence is the cumulative count of unique states visited over the course of a chain. At each step $t$, we record the number of distinct programs observed up to that point, as well as the number of distinct skeletons. By tracking both levels, we distinguish convergence in exact program identity from convergence in structural form. A chain may continue producing lexically distinct programs while visiting the same few skeletons, with variation limited to terminal substitutions (e.g., switching LEFT/RIGHT/FORWARD or swapping AND/OR) within a fixed structural template.

To characterize the structure of convergence, we also construct directed transition graphs from each chain, where nodes represent unique states (programs or skeletons) and directed edges represent observed mutations. We detect all simple cycles in these graphs and report their length distributions. Additionally, we compute the mean degree entropy across nodes, which captures how evenly transitions are distributed across successor states: low entropy indicates that each node tends to transition to a single successor, while higher entropy indicates more varied transitions.

As a complementary measure, we compute pairwise normalized token-level Levenshtein distances between successive programs, capturing the magnitude of change introduced by individual mutations.

\section{Experimental Design}

We conduct three experiments to assess the robustness of convergence dynamics under neutral LLM-driven mutation, varying prompt design, stochastic replication, and model family respectively. All experiments accept every valid mutation (one candidate per step, no selection pressure) with a chain length of 300 steps (accepted mutations). A classical GP subtree mutation baseline provides a point of comparison. Table~\ref{tab:params} summarizes the shared experimental parameters.

\begin{table}[h]
\caption{Shared experimental parameters. Experiment~3 uses different primary and fallback models for each condition; details are provided in Appendix~\ref{app:exps}.}
\begin{tabular}{ll}
\toprule
Parameter & Value \\
\midrule
Primary model & Gemini 3.1 Flash Lite \\
Fallback model & Gemini 3.1 Flash \\
Temperature & 1.0 \\
Max tokens & 1024 \\
Max retries (per model) & 5 \\
Chain length & 300 \\
\bottomrule
\end{tabular}
\label{tab:params}
\end{table}

\subsection{Experiment 1: Prompt Sensitivity}
\label{exp:prompt}

To assess the effect of prompt wording on trajectory dynamics, we sweep over 50 distinct prompt variants. Each prompt shares the same structure, with only the instruction line varying across conditions. The 50 unique instruction variants were generated by prompting ChatGPT to produce diverse paraphrasings of the core mutation command; the generation prompt and full list of variants are provided in Appendix~\ref{app:prompt-generation}. For each of the 50 prompts, we collect mutation chains starting from three different initial programs of varying size (small, medium, and large) drawn from a random initial population, yielding a total of 150 chains.

For comparison using the classical GP baseline, we also collect mutation chains using standard subtree mutation with a maximum depth of four, on the same three initial programs. We collect 50 independent chains per program, for a total of 150 baseline chains.

\subsection{Experiment 2: Intrinsic Variability}
\label{exp:traj}

To assess the stochastic variation inherent in the LLM mutation process itself, we run 30 independent trajectories for each of four prompts using the same medium-sized initial program and Gemini 3.1 Flash Lite, for a total of 120 chains. This experiment isolates the variability introduced by the model's sampling process, holding all other factors constant.

\subsection{Experiment 3: Model Sensitivity}
\label{exp:model}

To assess whether the observed convergence dynamics are model-specific or reflect a broader property of LLM-driven mutation, we run chains across seven models: Gemini 3.1 Flash Lite, Gemini 3.1 Flash Lite with reasoning, Claude Haiku 4.5, Claude Sonnet 4, Claude Sonnet 4.5, GPT-5 Mini, and GPT-5 Mini with reasoning. Each model is tested with four prompts and one initial program, for a total of 28 chains. Further details about the experimental parameters for each model are provided in Appendix~\ref{app:exps}.

\section{Results}
\label{sec:results}

Across all three experiments, LLM-driven mutation chains without selection pressure exhibit strong convergence, revisiting a small number of programs and structural forms rather than continuing to explore new regions of program space. The extent varies substantially with prompt wording and model choice, but the phenomenon is robust: the majority of chains visit far fewer unique programs and skeletons than the 300-step chain length would permit under standard GP subtree mutation.

We first examine sensitivity to prompt design (Section~\ref{res:promptsweep}), then assess stochastic consistency (Section~\ref{res:traj}), compare dynamics across seven LLMs (Section~\ref{res:model}), and analyze the cyclic structures underlying convergence (Section~\ref{res:cycle}).

\subsection{Experiment 1: Prompt Sensitivity}
\label{res:promptsweep}

\begin{figure}[!htbp]
  \centering
  \includegraphics[width=\linewidth]{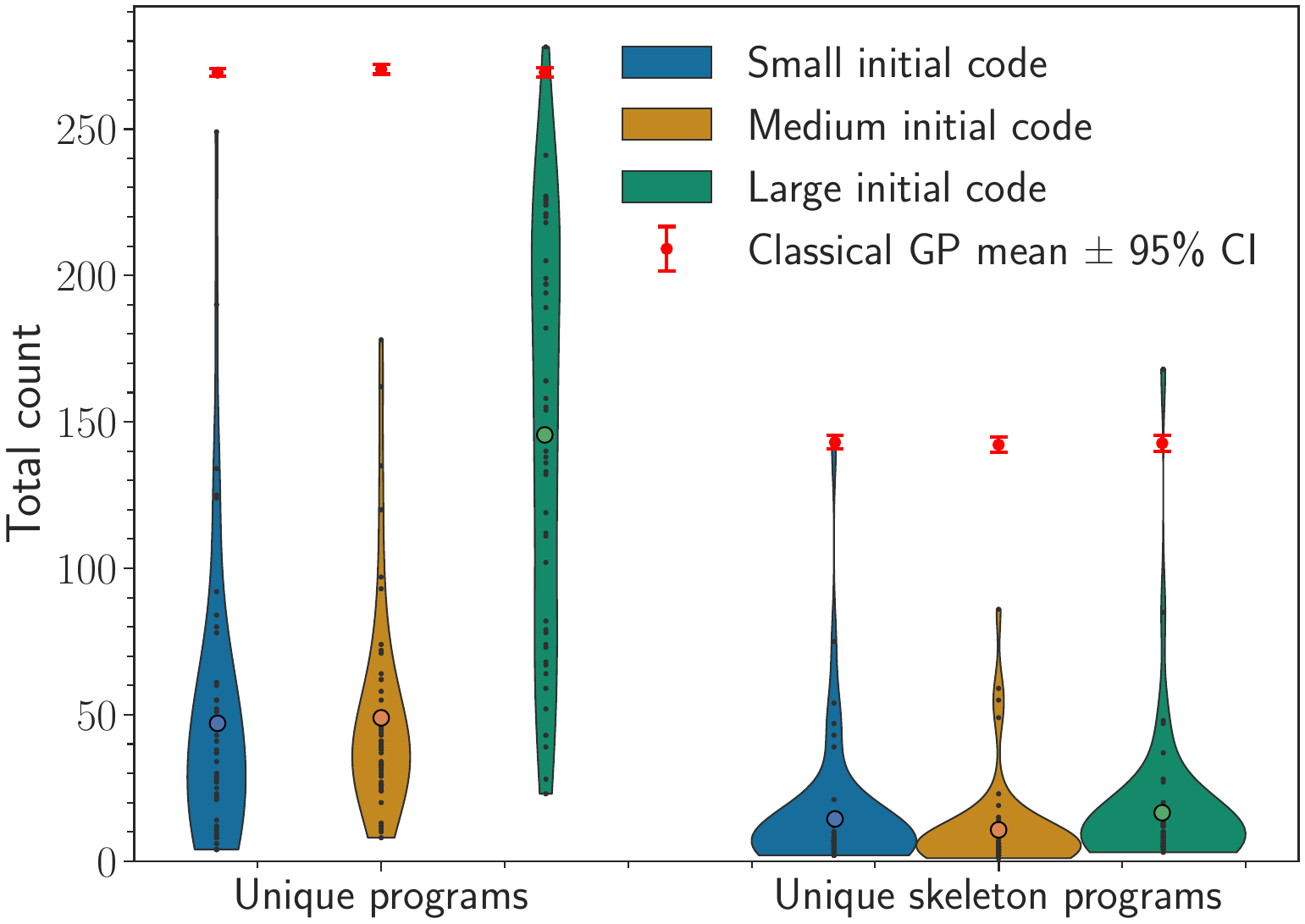}
  \caption{Distribution of cumulative unique programs and unique skeletons across 150 LLM mutation chains (50 prompts), grouped by three initial program sizes, with the classical GP subtree mutation baseline overlaid for comparison.}
  \Description{Bar or violin plots showing the distribution of cumulative unique programs and skeletons for LLM mutation chains grouped by initial program size. The GP baseline is overlaid, showing substantially higher unique counts than most LLM chains.}
  \label{fig:prompt-violin}
\end{figure}

Figure~\ref{fig:prompt-violin} summarizes the distribution of cumulative unique programs and unique skeletons across all 150 LLM mutation chains in the prompt sensitivity sweep, grouped by initial program size.

The wide variation in the resulting distributions demonstrates that the precise wording of the prompt plays a crucial role in the rate of convergence.  One of the most convergent prompts produced fewer than 10 unique programs across a 300-step chain, whereas one of the least convergent produced over 250, comparable to the classical GP subtree mutation baseline.  Initial program size, by contrast, has only a modest effect: chains initialized from the large program produce somewhat higher unique program counts, but the distributions across initialization sizes overlap substantially. 

At the skeleton level, convergence is more severe. The median LLM chain visits approximately 10 unique skeletons over 300 steps, indicating that for a wide range of prompts, LLM-generated variation consists primarily of terminal-level substitutions within a small set of recurring structural templates rather than exploration of substantially new program architectures.

The classical GP baseline (overlaid in Figure~\ref{fig:prompt-violin}) provides a direct comparison under identical conditions: the same DSL, the same initial programs, and the same chain length. GP chains visit approximately 270 unique programs and 143 unique skeletons per chain. This suggests that the size of the DSL is not a limiting factor: the space of reachable programs is large enough to sustain continued exploration under random subtree mutation. The convergence observed under LLM mutation reflects a property of the operator itself.

While the broader finding that prompt wording affects convergence might have been anticipated, two aspects of the results were unexpected:

\subparagraph{(i) The distribution of outcomes across prompts is heavily skewed:}  only a small fraction of the 50 prompts sustained substantial exploration, while the majority led to rapid convergence in the production of new genetic variants.

\subparagraph{(ii) The relationship between prompt wording and convergence behavior is difficult to predict:} prompts that appear semantically similar to one another can produce markedly different convergence profiles. See Appendix~\ref{app:prompt-sweep-results} for further examples.

Overall, of the 150 LLM-based mutation chains, 71\% visit fewer than 100 unique programs and 87\% visit fewer than 20 unique skeletons over 300 steps. Put differently, in 87\% of chains, over 93\% of individual mutations revisit a previously seen structural form. For most of the 50 prompts investigated, repeated LLM-based mutation tends to revisit the same programs (verbatim, or in common structural forms) far more frequently than classical GP subtree mutation.

\begin{figure}[!htbp]
  \centering
  \begin{subfigure}{0.49\columnwidth}
    \includegraphics[width=\textwidth]{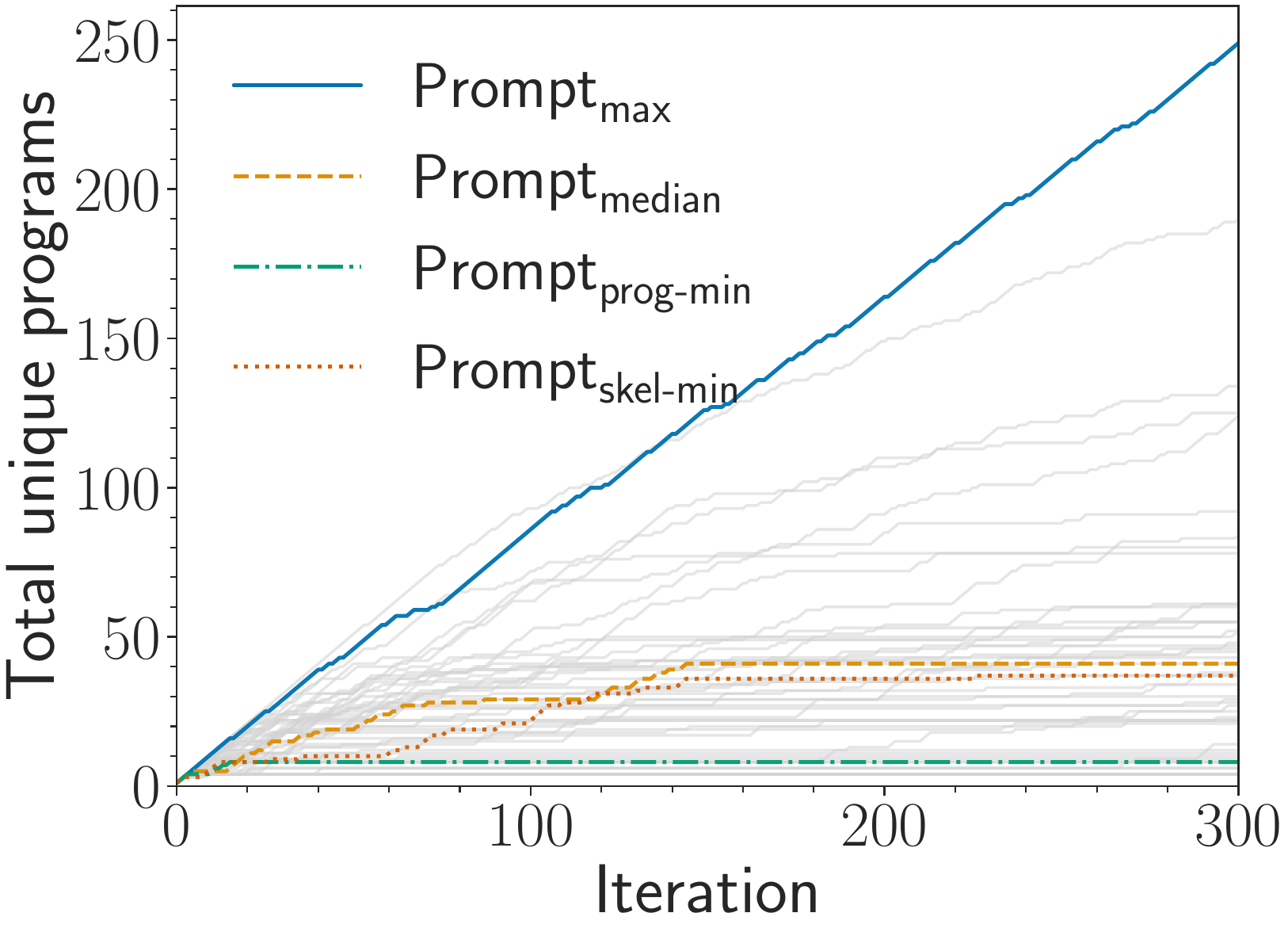}
    \caption{Programs}
  \end{subfigure}
   \begin{subfigure}{0.49\columnwidth}
    \includegraphics[width=\textwidth]{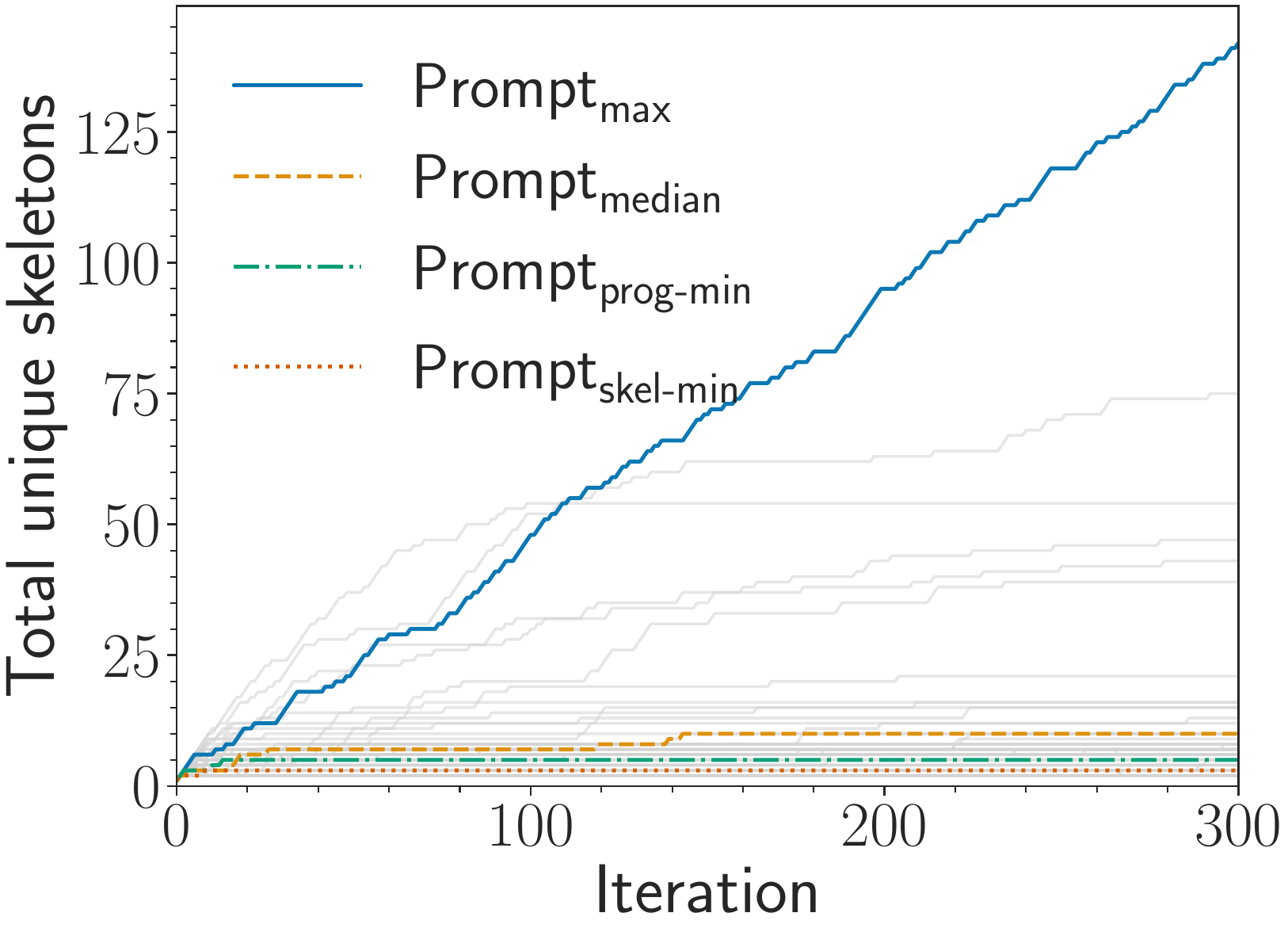}
    \caption{Skeletons}
  \end{subfigure}
  \caption{Cumulative unique programs (left) and unique skeletons (right) over 300 iterations for all 50 prompts (shown in faint gray) using the small initial program, with four representative prompts highlighted.}
  \Description{Two line plots showing cumulative unique counts over 300 iterations. Most lines flatten early, indicating convergence. Four highlighted prompts span the range from near-linear growth to rapid plateau.}
  \label{fig:prompt-traj}
\end{figure}

Figure~\ref{fig:prompt-traj} shows cumulative unique counts over 300 iterations for all 50 prompts, with four representative prompts highlighted. These four span the range of observed convergence behavior (Table~\ref{tab:prompts}): $\text{Prompt}_{\text{max}}$ (highest unique counts), $\text{Prompt}_{\text{median}}$ (median), $\text{Prompt}_{\text{prog-min}}$ (fewest unique programs), and $\text{Prompt}_{\text{skel-min}}$ (fewest unique skeletons). Because the latter two do not correlate, both are retained to capture distinct convergence dynamics. Most prompts show a characteristic flattening: unique counts rise in the first 50 to 100 steps before plateauing, while a small set (including $\text{Prompt}_{\text{max}}$) sustain near-linear growth throughout.

\begin{table}[h]
\caption{Selected prompt instructions used in Experiments 2 and 3.}
\begin{tabular}{p{0.25\columnwidth}p{0.65\columnwidth}}
\toprule
Label & Instruction \\
\midrule
$\text{Prompt}_{\text{max}}$ & \texttt{Generate a program that includes exploratory modifications.} \\
$\text{Prompt}_{\text{median}}$ & \texttt{Create a program with minor but meaningful modifications.} \\
$\text{Prompt}_{\text{prog-min}}$ & \texttt{Produce a program that has been slightly restructured.} \\
$\text{Prompt}_{\text{skel-min}}$ & \texttt{Generate a program with a small mutation applied.} \\
\bottomrule
\end{tabular}
\label{tab:prompts}
\end{table}

\subsection{Experiment 2: Intrinsic Variability}
\label{res:traj}

To assess whether the convergence patterns observed in Section~\ref{res:promptsweep} reflect stable properties of the prompt-model pairing or merely stochastic variation between runs, we replicate 30 independent chains for each of the four representative prompts identified in Table~\ref{tab:prompts}.

\begin{figure}[!htbp]
  \centering
  \begin{subfigure}{0.49\columnwidth}
    \includegraphics[width=\textwidth]{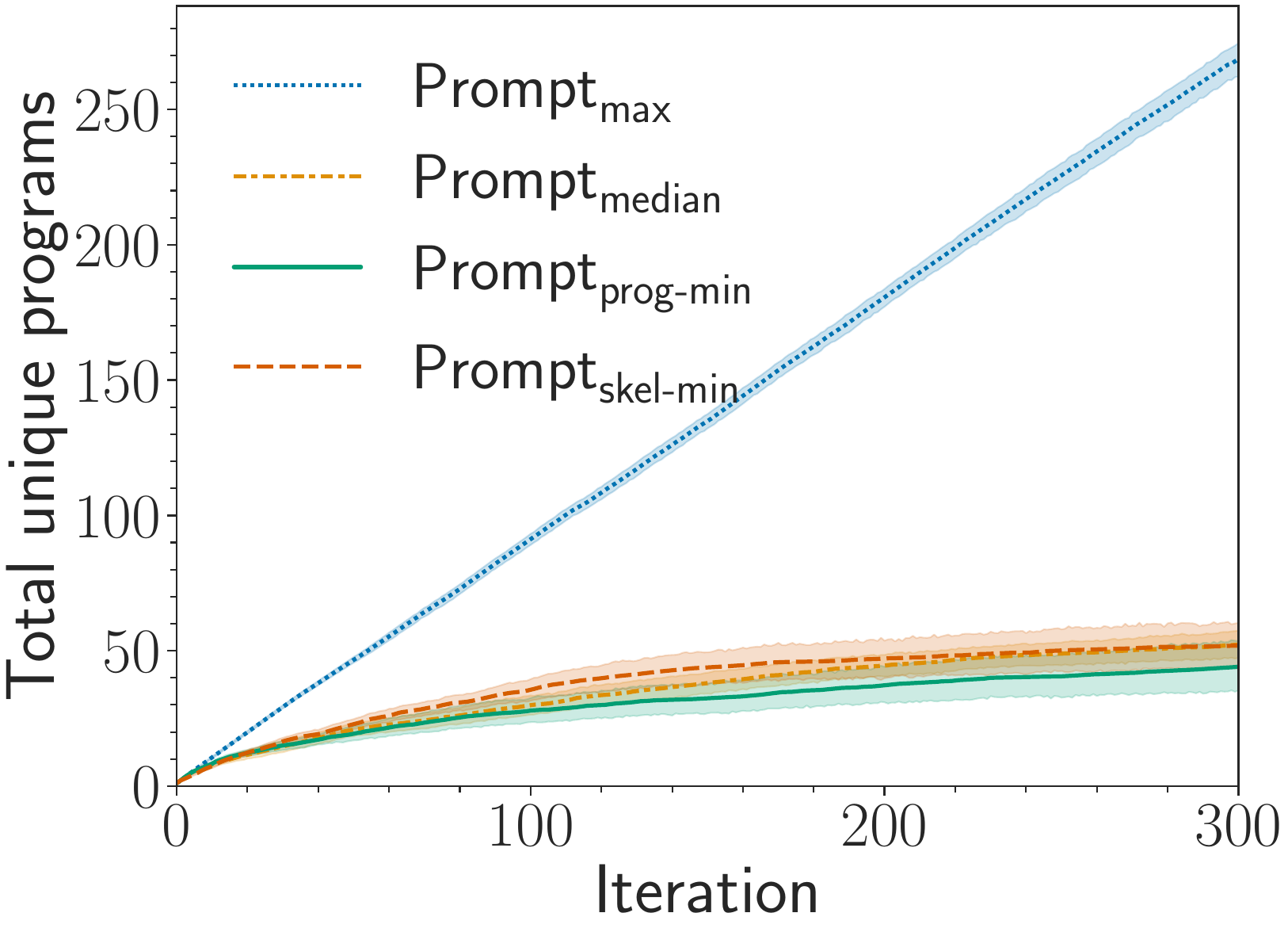}
    \caption{Programs}
     \label{fig:exp2-prog}
  \end{subfigure}
   \begin{subfigure}{0.49\columnwidth}
    \includegraphics[width=\textwidth]{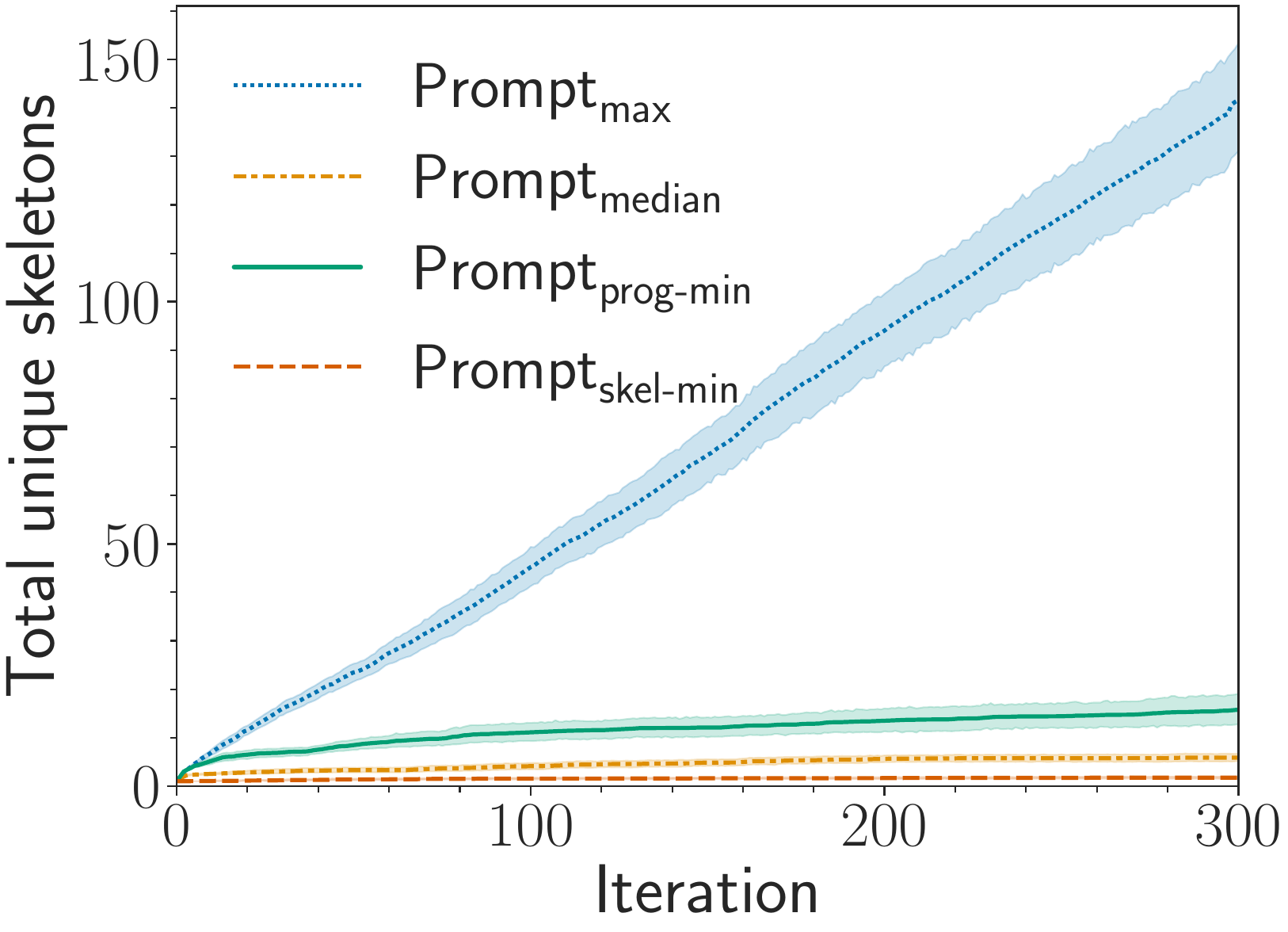}
    \caption{Skeletons}
     \label{fig:exp2-skel}
  \end{subfigure}
  \caption{Cumulative unique programs (left) and unique skeletons (right) over 300 iterations, averaged across 30 independent replications per prompt with 95\% confidence bands.}
  \Description{Two line plots with four curves each, showing mean cumulative unique counts with narrow confidence bands. Prompt max grows near-linearly while the other three prompts plateau within the first 50 to 100 steps.}
  \label{fig:exp2-stochasticity}
\end{figure}

Figure~\ref{fig:exp2-stochasticity} plots the cumulative unique program and skeleton counts over 300 iterations, averaged across the 30 replications with 95\% confidence bands. The trajectories separate into two distinct regimes. Under $\text{Prompt}_{\text{max}}$, unique program count grows near-linearly throughout the chain, reaching a mean of 269 unique programs and 140 unique skeletons by step 300. Under the remaining three prompts, the curves flatten within the first 50 to 100 steps: $\text{Prompt}_{\text{median}}$ reaches 52 unique programs and 6 unique skeletons, $\text{Prompt}_{\text{prog-min}}$ reaches 44 programs and 16 skeletons, and $\text{Prompt}_{\text{skel-min}}$ reaches 52 programs but only 1.7 unique skeletons.

The confidence bands are narrow relative to the separation between prompts, confirming that convergence behavior is highly consistent across independent runs of the same prompt. $\text{Prompt}_{\text{skel-min}}$ is a particularly striking case: across 30 independent chains, the mean number of unique skeletons visited is 1.7, meaning that most chains spend the entire 300-step trajectory cycling through variants of a single structural template.

\subsection{Experiment 3: Model Sensitivity}
\label{res:model}

\begin{figure}[!htbp]
  \centering
  \includegraphics[width=\linewidth]{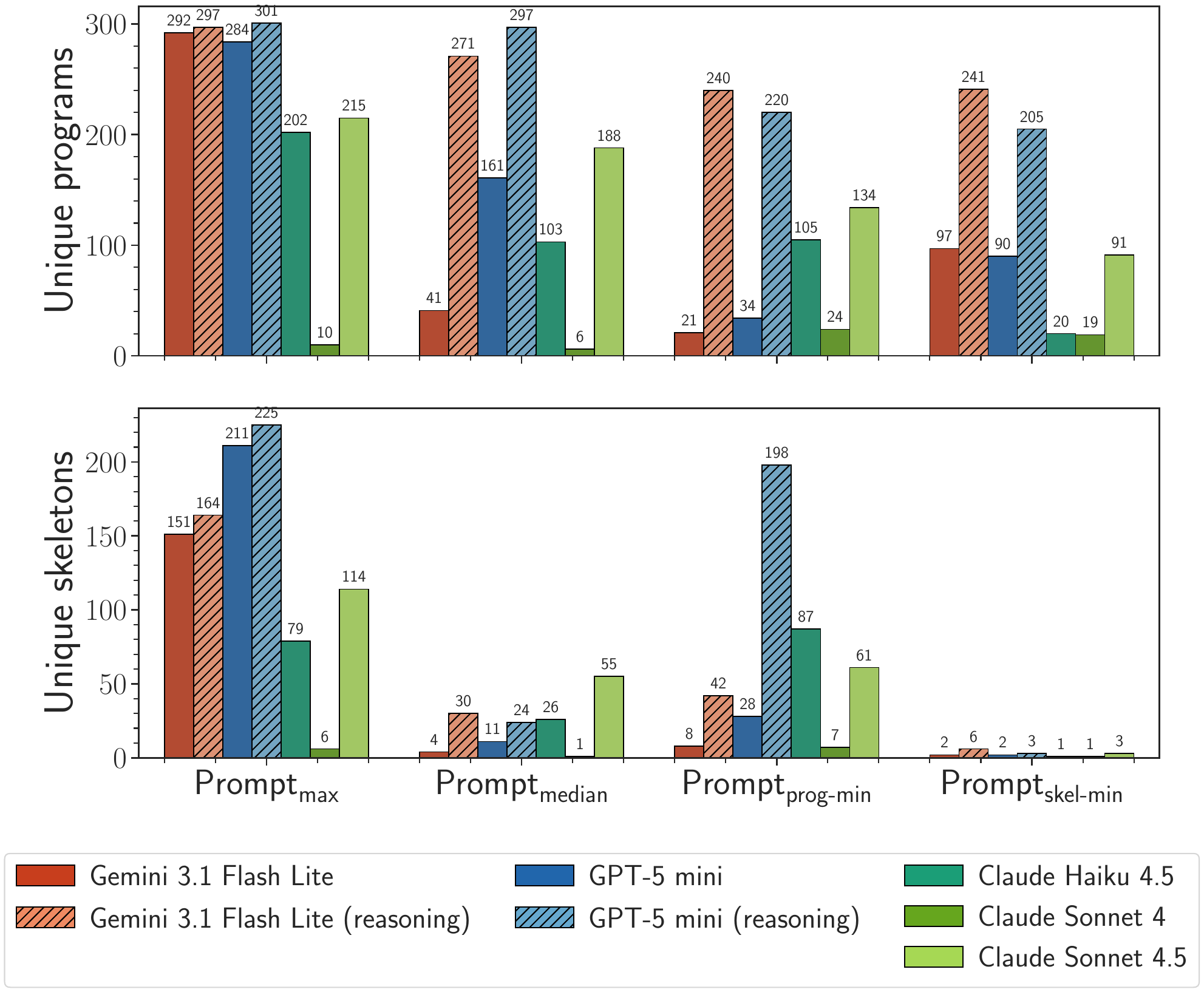}
  \caption{Cumulative unique programs and unique skeletons across seven LLMs, each tested with four representative prompts.}
  \Description{Grouped bar chart comparing cumulative unique programs and skeletons across seven models and four prompts. Models range from highly convergent (few unique counts) to highly exploratory (near 300 unique programs).}
  \label{fig:model-sweep}
\end{figure}

Figure~\ref{fig:model-sweep} compares cumulative unique programs and skeletons across seven LLMs, each tested with the same four prompts from Table~\ref{tab:prompts}. The results reveal substantial variation across model families that persists regardless of prompt.

At one extreme, Claude Sonnet 4 produced as few as 6 unique programs and 1 unique skeleton over 300 steps, collapsing to a single structural form almost immediately regardless of prompt. At the other extreme, GPT-5 Mini with reasoning produced up to 301 unique programs and 225 unique skeletons, sustaining near-complete exploration even under the prompt that produced the strongest convergence in the Gemini-based prompt sweep. 

While some models sustain greater program diversity, the majority of the seven models tested still exhibit convergent behavior across most or all prompts. Two further patterns emerge. First, the ranking is largely prompt-invariant: models that converge under $\text{Prompt}_{\text{median}}$ also converge under $\text{Prompt}_{\text{max}}$, and models that maintain a high rate of exploration under $\text{Prompt}_{\text{max}}$ also explore under more convergent prompts. This suggests that the tendency to converge reflects a persistent property of the model rather than an interaction with specific prompt wording. Second, reasoning-enabled variants consistently produce higher diversity than their base counterparts: GPT-5 Mini with reasoning substantially outperforms GPT-5 Mini, and Gemini 3.1 Flash Lite with reasoning outperforms the base Gemini model across all four prompts. Whether this reflects the additional computation afforded by reasoning or a difference in how these models represent and transform code is beyond the scope of this study.

\subsection{Attractor Structure and Cycle Analysis}
\label{res:cycle}

The preceding sections establish that, across a range of prompts, initial programs, and models, LLM mutation chains tend to ``converge'', visiting a restricted set of programs and skeletons. To characterize the behavior of this convergence, we construct directed transition graphs inspired by~\cite{vanstein2025codeevolgraphs}, where nodes are unique states and directed edges represent observed mutations (an example is shown in Figure~\ref{fig:sample-mutation-trajectory}). We build these graphs from the 30-replication experiment (Section~\ref{res:traj}) and analyze the cycles within them.

Table~\ref{tab:graph-program} reports graph-level statistics at the program and skeleton level, averaged over 30 replications per prompt condition.

\begin{table*}[h]
\caption{Program and skeleton-level transition graph statistics, averaged over 30 replications per prompt condition. Values are reported as mean $\pm$ standard deviation.}
\label{tab:graph-program}
\begin{tabular}{clcccc}
\toprule
Graph Type & Prompt & Mean Cycle Length & Min Cycle Length & Max Cycle Length & Mean Degree Entropy \\
\midrule
\multirow{4}{5em}{Program} 
& $\text{Prompt}_{\text{skel-min}}$  & $3.41 \pm 1.92$ & $2.00 \pm 0.00$ & $8.87 \pm 5.73$ & $0.085 \pm 0.019$ \\
& $\text{Prompt}_{\text{prog-min}}$  & $3.42 \pm 1.65$ & $1.77 \pm 0.43$ & $8.00 \pm 5.13$ & $0.104 \pm 0.032$ \\
& $\text{Prompt}_{\text{median}}$       & $5.60 \pm 2.63$ & $2.00 \pm 0.00$ & $13.63 \pm 5.49$ & $0.091 \pm 0.014$ \\
& $\text{Prompt}_{\text{max}}$          & $4.03 \pm 4.34$ & $1.03 \pm 0.18$ & $21.33 \pm 32.28$ & $0.030 \pm 0.001$ \\
\midrule
\multirow{4}{5em}{Skeleton} 
& $\text{Prompt}_{\text{skel-min}}$  & $1.04 \pm 0.09$ & $1.00 \pm 0.00$ & $1.13 \pm 0.35$ & $0.226 \pm 0.198$ \\
& $\text{Prompt}_{\text{prog-min}}$  & $1.71 \pm 0.53$ & $1.00 \pm 0.00$ & $3.90 \pm 2.32$ & $0.122 \pm 0.036$ \\
& $\text{Prompt}_{\text{median}}$       & $1.10 \pm 0.15$ & $1.00 \pm 0.00$ & $1.50 \pm 0.68$ & $0.173 \pm 0.076$ \\
& $\text{Prompt}_{\text{max}}$          & $3.54 \pm 3.28$ & $1.00 \pm 0.00$ & $32.50 \pm 25.36$ & $0.049 \pm 0.009$ \\
\bottomrule
\end{tabular}
\end{table*}

\begin{figure}[!htbp]
  \centering
  \begin{subfigure}{0.49\columnwidth}
    \includegraphics[width=\textwidth]{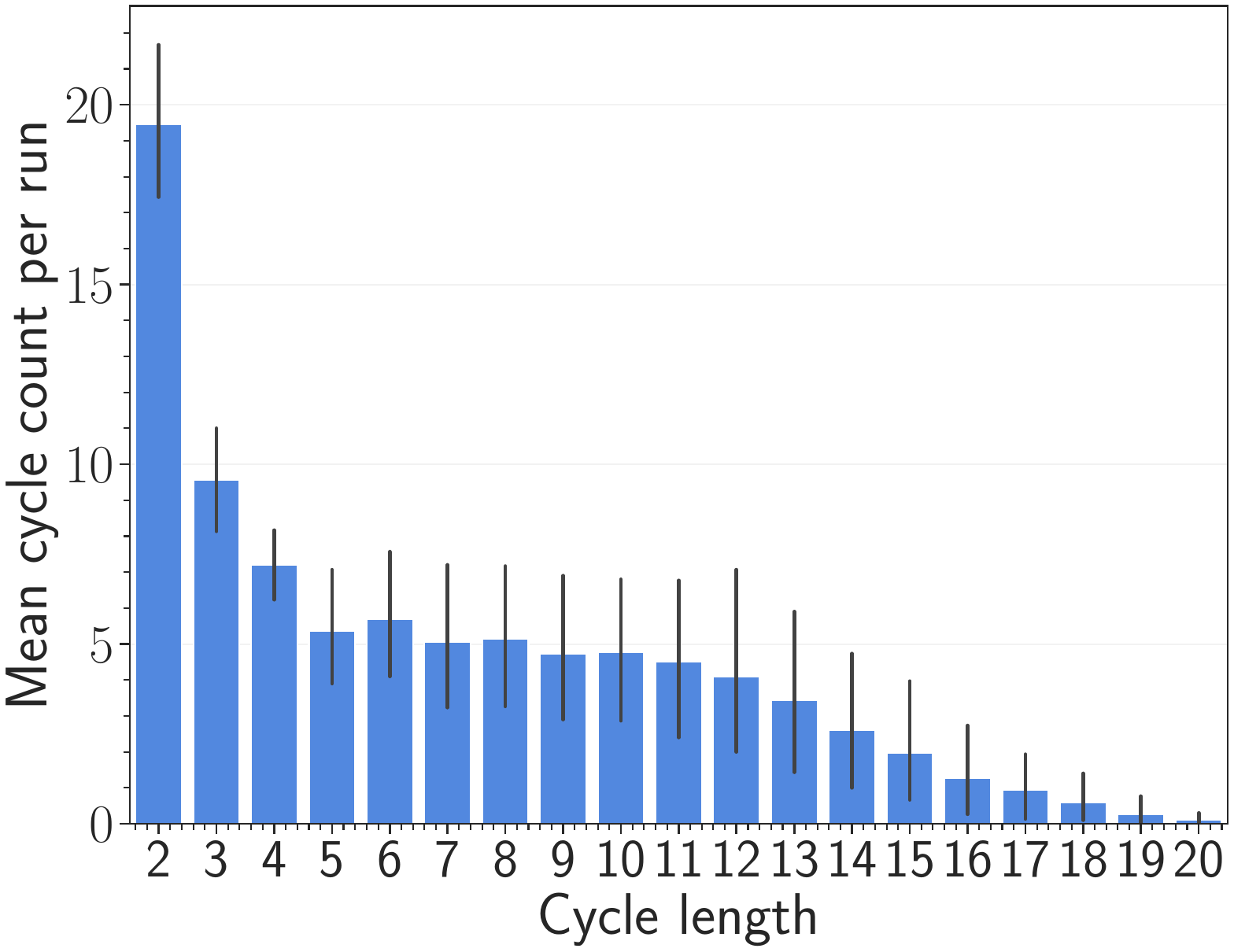}
    \caption{Programs}
  \end{subfigure}
   \begin{subfigure}{0.49\columnwidth}
    \includegraphics[width=\textwidth]{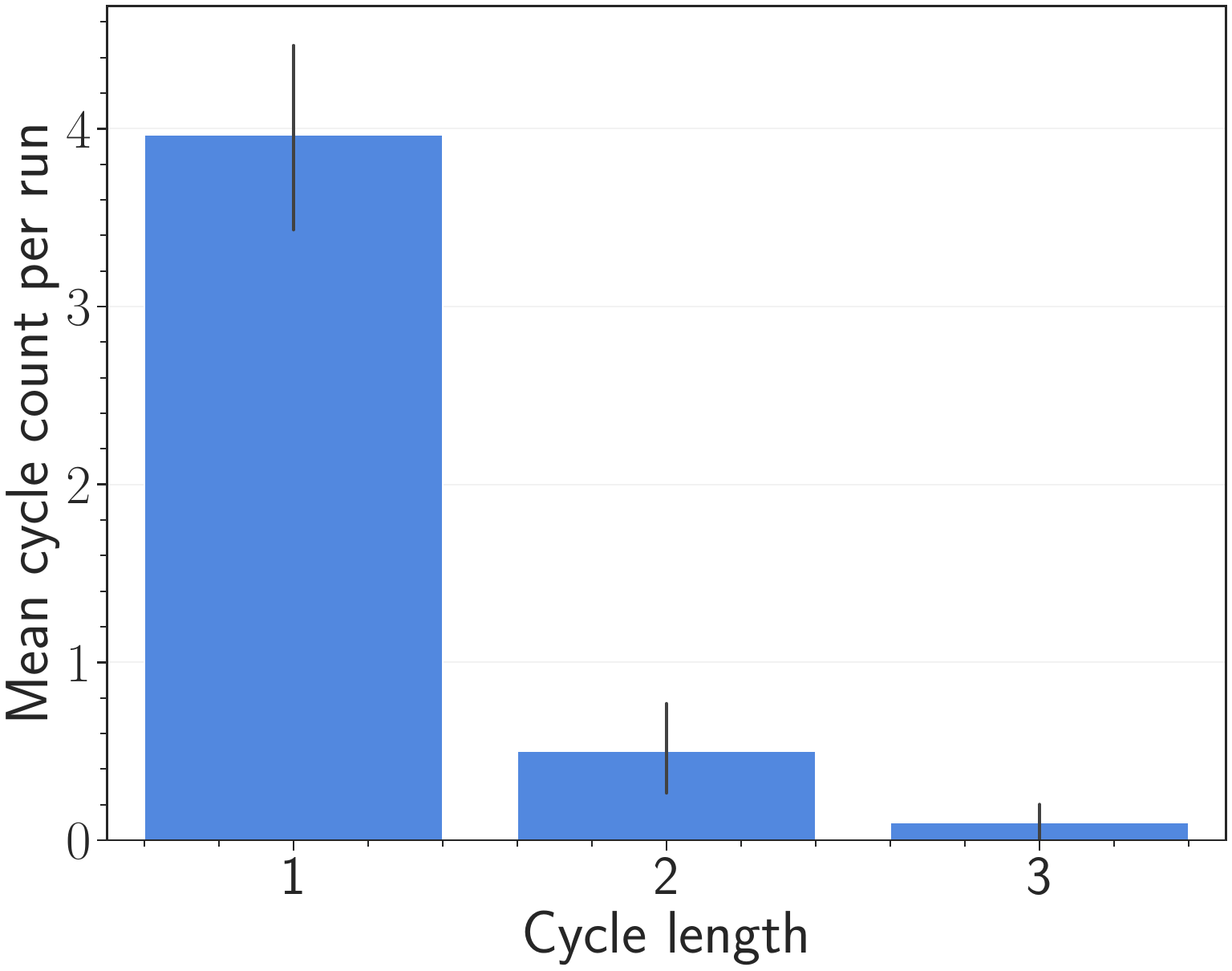}
    \caption{Skeletons}
  \end{subfigure}
  \caption{Cycle length distributions for $\text{Prompt}_{\text{median}}$ at the program level (left) and skeleton level (right), averaged over 30 replications.}
  \Description{Two histograms of cycle lengths. The program-level histogram shows a heavy-tailed distribution dominated by cycles of length 2 and 3. The skeleton-level histogram is concentrated almost entirely at length 1.}
  \label{fig:cycle-hist}
\end{figure}

Short cycles dominate at both levels. At the program level, the minimum cycle length is 2 for three of the four prompts, indicating that 2-cycles are a consistent feature of the transition structure. Mean cycle lengths range from 3.4 to 5.6 across conditions. At the skeleton level, the distribution is more extreme: under the more convergent prompts ($\text{Prompt}_{\text{skel-min}}$, $\text{Prompt}_{\text{median}}$), mean cycle lengths are close to 1 (1.04 and 1.10 respectively), meaning nearly all skeleton-level cycles are self-loops. Figure~\ref{fig:cycle-hist} shows these distributions for $\text{Prompt}_{\text{median}}$.

Degree entropy captures how evenly transitions are distributed across successor states. Convergent prompts show higher program-level degree entropy (0.085--0.104) than $\text{Prompt}_{\text{max}}$ (0.030), reflecting the fact that chains revisiting a small set of nodes accumulate multiple outgoing edges per node, while exploratory chains visit most nodes only once.

$\text{Prompt}_{\text{max}}$ is a notable exception at the skeleton level, averaging 73 cycles with a mean length of 3.5 and a maximum of 32.5. This indicates that even chains sustaining structural exploration develop recurrent patterns at longer scales.

Figure~\ref{fig:sample-mutation-trajectory} shows a representative program-level transition graph from a single chain using $\text{Prompt}_{\text{median}}$ and the medium initial program ($C_{\text{medium}}$). The trajectory passes through a transient sequence of states before settling into a 2-cycle: two program states that alternate indefinitely, with the majority of the 300 steps spent oscillating between them. This 2-cycle structure is consistent with the minimum cycle lengths reported in Table~\ref{tab:graph-program} and parallels the 2-period attractor cycles identified by Wang et al.~\cite{wang2025attractorcycles} in successive paraphrasing. The corresponding skeleton-level transition graph for this same run is provided in Appendix~\ref{sec:additional-results}. Pairwise normalized Levenshtein distance heatmaps, which visualize the temporal fine structure of these attractor dynamics at the individual-chain level, are presented in Appendix~\ref{sec:heatmaps}.

\section{Discussion}

The results provide clear answers to each of the three research questions posed in Section~\ref{sec:rqs}. First, LLM-driven mutation chains do converge toward attractor regions: the majority of chains plateau in their cumulative unique counts within the first 50 to 100 steps, and cycle analysis reveals structured revisitation patterns, with 2-cycles dominating at the program level and self-loops dominating at the skeleton level. Second, diversity stagnation is pronounced and operates at two distinct levels: while chains may continue producing lexically distinct programs (especially when mutating larger programs), structural diversity (measured at the skeleton level) stagnates far more severely, with some chains visiting only one or two unique skeletons across 300 steps. Third, these dynamics are sensitive to both prompt design and model choice but robust in aggregate: convergence is the default outcome across the majority of prompts and models tested, with prompt wording and model family modulating the severity but not eliminating the phenomenon.

These findings align with and extend convergence results reported in other iterative LLM domains. The dominance of short cycles in program-level transition graphs parallels the 2-period limit cycles identified by Wang et al.~\cite{wang2025attractorcycles} in successive paraphrasing. The collapse toward a small number of recurring structural forms is analogous to the 12 dominant visual motifs observed by Hintze et al.~\cite{hintze2025imagemotifs} in text-to-image-to-text loops. That these patterns emerge across text, image, and now program space suggests that convergence under iterative application may be a general property of current LLM architectures rather than a domain-specific artifact. Our results also complement the findings of the Digital Red Queen study~\cite{digitalredqueen}, which reported convergence in behavior space but not in syntactic program space under adversarial co-evolution with selection pressure. In our setting, without selection pressure, convergence manifests directly in the program space itself, suggesting that in some instances selection may mask rather than prevent structural convergence.

For practitioners building systems that employ LLMs as code mutation or generation operators, these results warrant careful attention. The tendency of LLM mutation operators toward structural convergence is more pronounced for smaller models and for models without reasoning capabilities, but it is difficult to predict in advance how a given prompt-model pairing will affect the search space. The finding that prompt design substantially modulates convergence severity is encouraging, but the relationship between prompt wording and exploration behavior can be unintuitive: prompts that appear semantically similar can produce very different convergence profiles. Practitioners may therefore benefit from conducting mutation-chain analysis in the absence of selection pressure to identify prompts that sustain exploration. More broadly, systems that require sustained structural diversity may benefit from explicit diversity maintenance mechanisms, as several recent systems have already recognized. Selection pressure itself may also help drive the LLM beyond attractors in program space, though the interaction between selection and the operator's intrinsic convergence dynamics remains an open question.

Several limitations should be noted. The experiments use a single constrained Lisp-like DSL, and the extent to which these dynamics generalize to richer programming languages, which are more strongly represented in LLM pretraining data, remains an open question. The analysis is purely genotypic: programs that are structurally identical may differ in behavior, and programs that converge structurally may still provide useful behavioral variation. Due to computational constraints, the model sensitivity experiment used only four prompts per model with no replication, limiting the strength of model-level claims. Additionally, for models with high retry rates, the error-correction process may contribute to observed variation, though retries correct syntax errors rather than produce independent mutations.

Future work should extend this analysis to broader program representations and incorporate behavioral evaluation. A particularly promising direction is investigating how diversity preservation mechanisms and selection pressure interact with the operator's intrinsic convergence dynamics, and whether such mechanisms can effectively drive LLM-based mutation beyond the attractor regions identified here. Further formalization of the dynamical systems framework, including characterization of attractor basins, contraction rates, and embedding-space representations of mutation trajectories, may deepen our understanding of the mechanisms driving convergence.

\section{Conclusion}

This work provides a systematic empirical investigation of LLM-driven mutation as a dynamical process over program space. By studying mutation chains in the absence of selection pressure across 50 prompt variants, seven models, and 30-fold replications, we demonstrate that convergence toward restricted structural regions is a robust property of LLM-based mutation operators in this constrained DSL setting. The rate and severity of convergence depend on prompt design and model choice, but the phenomenon itself is pervasive. These findings have direct implications for any system that relies on iterative LLM-driven code transformation: the mutation operator that enables semantics-aware program transformation also carries a systematic bias toward structural homogeneity that must be understood and accounted for if LLM-driven systems are to sustain the open-ended exploration on which evolutionary computation, automated scientific discovery, and iterative program synthesis increasingly depend.

\begin{acks}
The authors used generative AI tools for coding assistance and to improve the readability of this manuscript.
\end{acks}

\bibliographystyle{ACM-Reference-Format}
\bibliography{lmca_refs}


\begin{thebibliography}{50}


\ifx \showCODEN    \undefined \def \showCODEN     #1{\unskip}     \fi
\ifx \showISBNx    \undefined \def \showISBNx     #1{\unskip}     \fi
\ifx \showISBNxiii \undefined \def \showISBNxiii  #1{\unskip}     \fi
\ifx \showISSN     \undefined \def \showISSN      #1{\unskip}     \fi
\ifx \showLCCN     \undefined \def \showLCCN      #1{\unskip}     \fi
\ifx \shownote     \undefined \def \shownote      #1{#1}          \fi
\ifx \showarticletitle \undefined \def \showarticletitle #1{#1}   \fi
\ifx \showURL      \undefined \def \showURL       {\relax}        \fi
\providecommand\bibfield[2]{#2}
\providecommand\bibinfo[2]{#2}
\providecommand\natexlab[1]{#1}
\providecommand\showeprint[2][]{arXiv:#2}

\bibitem[Atkinson et~al\mbox{.}(2018)]%
        {atkinson2018snd}
\bibfield{author}{\bibinfo{person}{Timothy Atkinson}, \bibinfo{person}{Detlef
  Plump}, {and} \bibinfo{person}{Susan Stepney}.}
  \bibinfo{year}{2018}\natexlab{}.
\newblock \showarticletitle{Evolving Graphs by Graph Programming}. In
  \bibinfo{booktitle}{\emph{Genetic Programming}},
  \bibfield{editor}{\bibinfo{person}{Mauro Castelli}, \bibinfo{person}{Lukas
  Sekanina}, \bibinfo{person}{Mengjie Zhang}, \bibinfo{person}{Stefano
  Cagnoni}, {and} \bibinfo{person}{Pablo Garc{\'i}a-S{\'a}nchez}} (Eds.).
  \bibinfo{publisher}{Springer International Publishing},
  \bibinfo{address}{Cham}, \bibinfo{pages}{35--51}.
\newblock
\showISBNx{978-3-319-77553-1}


\bibitem[Banzhaf et~al\mbox{.}(2006)]%
        {EA-Banzhaf}
\bibfield{author}{\bibinfo{person}{Wolfgang Banzhaf},
  \bibinfo{person}{Guillaume Beslon}, \bibinfo{person}{Steffen Christensen},
  \bibinfo{person}{James~A. Foster}, \bibinfo{person}{François Képès},
  \bibinfo{person}{Virginie Lefort}, \bibinfo{person}{Julian~F. Miller},
  \bibinfo{person}{Miroslav Radman}, {and} \bibinfo{person}{Jeremy~J.
  Ramsden}.} \bibinfo{year}{2006}\natexlab{}.
\newblock \showarticletitle{Guidelines: From artificial evolution to
  computational evolution: A research agenda}.
\newblock \bibinfo{journal}{\emph{Nature Reviews Genetics}}
  \bibinfo{volume}{7} (\bibinfo{year}{2006}).
\newblock
Issue 9.
\showISSN{14710056}
\href{https://doi.org/10.1038/nrg1921}{doi:\nolinkurl{10.1038/nrg1921}}


\bibitem[Bradley et~al\mbox{.}(2024)]%
        {LLM_OpenELM}
\bibfield{author}{\bibinfo{person}{Herbie Bradley}, \bibinfo{person}{Honglu
  Fan}, \bibinfo{person}{Theodoros Galanos}, \bibinfo{person}{Ryan Zhou},
  \bibinfo{person}{Daniel Scott}, {and} \bibinfo{person}{Joel Lehman}.}
  \bibinfo{year}{2024}\natexlab{}.
\newblock \bibinfo{booktitle}{\emph{The OpenELM Library: Leveraging Progress in
  Language Models for Novel Evolutionary Algorithms}}.
\newblock \bibinfo{publisher}{Springer Nature Singapore},
  \bibinfo{address}{Singapore}, \bibinfo{pages}{177--201}.
\newblock
\showISBNx{978-981-99-8413-8}
\href{https://doi.org/10.1007/978-981-99-8413-8_10}{doi:\nolinkurl{10.1007/978-981-99-8413-8_10}}


\bibitem[Custode et~al\mbox{.}(2024)]%
        {LLM_hyperparam_EA}
\bibfield{author}{\bibinfo{person}{Leonardo~Lucio Custode},
  \bibinfo{person}{Fabio Caraffini}, \bibinfo{person}{Anil Yaman}, {and}
  \bibinfo{person}{Giovanni Iacca}.} \bibinfo{year}{2024}\natexlab{}.
\newblock \showarticletitle{An investigation on the use of Large Language
  Models for hyperparameter tuning in Evolutionary Algorithms}. In
  \bibinfo{booktitle}{\emph{Proceedings of the Genetic and Evolutionary
  Computation Conference Companion}} (Melbourne, VIC, Australia)
  \emph{(\bibinfo{series}{GECCO '24 Companion})}.
  \bibinfo{publisher}{Association for Computing Machinery},
  \bibinfo{address}{New York, NY, USA}, \bibinfo{pages}{1838–1845}.
\newblock
\showISBNx{9798400704956}
\href{https://doi.org/10.1145/3638530.3664163}{doi:\nolinkurl{10.1145/3638530.3664163}}


\bibitem[Darwin(1859)]%
        {darwin1859}
\bibfield{author}{\bibinfo{person}{Charles Darwin}.}
  \bibinfo{year}{1859}\natexlab{}.
\newblock \bibinfo{booktitle}{\emph{On the Origin of Species by Means of
  Natural Selection}}.
\newblock \bibinfo{publisher}{Murray}, \bibinfo{address}{London}.
\newblock
\newblock
\shownote{or the Preservation of Favored Races in the Struggle for Life}.


\bibitem[Fernando et~al\mbox{.}(2024)]%
        {llm_promptbreeder}
\bibfield{author}{\bibinfo{person}{Chrisantha Fernando},
  \bibinfo{person}{Dylan~Sunil Banarse}, \bibinfo{person}{Henryk Michalewski},
  \bibinfo{person}{Simon Osindero}, {and} \bibinfo{person}{Tim
  Rockt\"{a}schel}.} \bibinfo{year}{2024}\natexlab{}.
\newblock \showarticletitle{Promptbreeder: Self-Referential Self-Improvement
  via Prompt Evolution}. In \bibinfo{booktitle}{\emph{Proceedings of the 41st
  International Conference on Machine Learning}}
  \emph{(\bibinfo{series}{Proceedings of Machine Learning Research},
  Vol.~\bibinfo{volume}{235})}, \bibfield{editor}{\bibinfo{person}{Ruslan
  Salakhutdinov}, \bibinfo{person}{Zico Kolter}, \bibinfo{person}{Katherine
  Heller}, \bibinfo{person}{Adrian Weller}, \bibinfo{person}{Nuria Oliver},
  \bibinfo{person}{Jonathan Scarlett}, {and} \bibinfo{person}{Felix
  Berkenkamp}} (Eds.). \bibinfo{publisher}{PMLR},
  \bibinfo{pages}{13481--13544}.
\newblock
\urldef\tempurl%
\url{https://proceedings.mlr.press/v235/fernando24a.html}
\showURL{%
\tempurl}


\bibitem[Guo et~al\mbox{.}(2026)]%
        {guo2026codeevol}
\bibfield{author}{\bibinfo{person}{Ping Guo}, \bibinfo{person}{Chao Li},
  \bibinfo{person}{Yinglan Feng}, {and} \bibinfo{person}{Chaoning Zhang}.}
  \bibinfo{year}{2026}\natexlab{}.
\newblock \bibinfo{title}{Code Evolution for Control: Synthesizing Policies via
  LLM-Driven Evolutionary Search}.
\newblock
\showeprint[arxiv]{2601.06845}~[cs.AI]
\urldef\tempurl%
\url{https://arxiv.org/abs/2601.06845}
\showURL{%
\tempurl}


\bibitem[Gurkan et~al\mbox{.}(2025)]%
        {LEAR}
\bibfield{author}{\bibinfo{person}{Can Gurkan},
  \bibinfo{person}{Narasimha~Karthik Jwalapuram}, \bibinfo{person}{Kevin Wang},
  \bibinfo{person}{Rudy Danda}, \bibinfo{person}{Leif Rasmussen},
  \bibinfo{person}{John Chen}, {and} \bibinfo{person}{Uri Wilensky}.}
  \bibinfo{year}{2025}\natexlab{}.
\newblock \showarticletitle{LEAR: LLM-Driven Evolution of Agent-Based Rules}.
  In \bibinfo{booktitle}{\emph{Proceedings of the Genetic and Evolutionary
  Computation Conference Companion}} (NH Malaga Hotel, Malaga, Spain)
  \emph{(\bibinfo{series}{GECCO '25 Companion})}.
  \bibinfo{publisher}{Association for Computing Machinery},
  \bibinfo{address}{New York, NY, USA}, \bibinfo{pages}{2309–2326}.
\newblock
\showISBNx{9798400714641}
\href{https://doi.org/10.1145/3712255.3734368}{doi:\nolinkurl{10.1145/3712255.3734368}}


\bibitem[Hagos et~al\mbox{.}(2024)]%
        {LLM_recent_advances}
\bibfield{author}{\bibinfo{person}{Desta~Haileselassie Hagos},
  \bibinfo{person}{Rick Battle}, {and} \bibinfo{person}{Danda~B. Rawat}.}
  \bibinfo{year}{2024}\natexlab{}.
\newblock \showarticletitle{Recent Advances in Generative AI and Large Language
  Models: Current Status, Challenges, and Perspectives}.
\newblock \bibinfo{journal}{\emph{IEEE Transactions on Artificial
  Intelligence}} \bibinfo{volume}{5}, \bibinfo{number}{12}
  (\bibinfo{year}{2024}), \bibinfo{pages}{5873--5893}.
\newblock
\href{https://doi.org/10.1109/TAI.2024.3444742}{doi:\nolinkurl{10.1109/TAI.2024.3444742}}


\bibitem[Hemberg et~al\mbox{.}(2025)]%
        {LLM_GP_review}
\bibfield{author}{\bibinfo{person}{Erik Hemberg}, \bibinfo{person}{Steven
  Jorgensen}, {and} \bibinfo{person}{Una-May O'Reilly}.}
  \bibinfo{year}{2025}\natexlab{}.
\newblock \bibinfo{booktitle}{\emph{Survey of Genetic Programming and Large
  Language Models}}.
\newblock \bibinfo{publisher}{Springer Nature Singapore},
  \bibinfo{address}{Singapore}, \bibinfo{pages}{67--86}.
\newblock
\showISBNx{978-981-96-0077-9}
\href{https://doi.org/10.1007/978-981-96-0077-9_4}{doi:\nolinkurl{10.1007/978-981-96-0077-9_4}}


\bibitem[Hemberg et~al\mbox{.}(2024)]%
        {LLM_GP_code_Hemberg2024}
\bibfield{author}{\bibinfo{person}{Erik Hemberg}, \bibinfo{person}{Stephen
  Moskal}, {and} \bibinfo{person}{Una-May O'Reilly}.}
  \bibinfo{year}{2024}\natexlab{}.
\newblock \showarticletitle{Evolving code with a large language model}.
\newblock \bibinfo{journal}{\emph{Genetic Programming and Evolvable Machines}}
  \bibinfo{volume}{25}, \bibinfo{number}{2} (\bibinfo{date}{12 Sep}
  \bibinfo{year}{2024}), \bibinfo{pages}{21}.
\newblock
\showISSN{1573-7632}
\href{https://doi.org/10.1007/s10710-024-09494-2}{doi:\nolinkurl{10.1007/s10710-024-09494-2}}


\bibitem[Hintze et~al\mbox{.}(2026)]%
        {hintze2025imagemotifs}
\bibfield{author}{\bibinfo{person}{Arend Hintze}, \bibinfo{person}{Frida
  Proschinger~{\AA}str{\"o}m}, {and} \bibinfo{person}{Jory Schossau}.}
  \bibinfo{year}{2026}\natexlab{}.
\newblock \showarticletitle{Autonomous language-image generation loops converge
  to generic visual motifs}.
\newblock \bibinfo{journal}{\emph{Patterns}} \bibinfo{volume}{7},
  \bibinfo{number}{1} (\bibinfo{date}{09 Jan} \bibinfo{year}{2026}).
\newblock
\showISSN{2666-3899}
\href{https://doi.org/10.1016/j.patter.2025.101451}{doi:\nolinkurl{10.1016/j.patter.2025.101451}}


\bibitem[Hu and Zhang(2025)]%
        {partevo}
\bibfield{author}{\bibinfo{person}{Qinglong Hu} {and} \bibinfo{person}{Qingfu
  Zhang}.} \bibinfo{year}{2025}\natexlab{}.
\newblock \showarticletitle{Partition to Evolve: Niching-enhanced Evolution
  with {LLM}s for Automated Algorithm Discovery}. In
  \bibinfo{booktitle}{\emph{The Thirty-ninth Annual Conference on Neural
  Information Processing Systems}}.
\newblock
\urldef\tempurl%
\url{https://openreview.net/forum?id=OEawM2coNT}
\showURL{%
\tempurl}


\bibitem[Hu et~al\mbox{.}(2025)]%
        {LLM_ADAS}
\bibfield{author}{\bibinfo{person}{Shengran Hu}, \bibinfo{person}{Cong Lu},
  {and} \bibinfo{person}{Jeff Clune}.} \bibinfo{year}{2025}\natexlab{}.
\newblock \showarticletitle{Automated Design of Agentic Systems}. In
  \bibinfo{booktitle}{\emph{The Thirteenth International Conference on Learning
  Representations}}.
\newblock
\urldef\tempurl%
\url{https://openreview.net/forum?id=t9U3LW7JVX}
\showURL{%
\tempurl}


\bibitem[Jiang et~al\mbox{.}(2024)]%
        {LLM_survey_code}
\bibfield{author}{\bibinfo{person}{Juyong Jiang}, \bibinfo{person}{Fan Wang},
  \bibinfo{person}{Jiasi Shen}, \bibinfo{person}{Sungju Kim}, {and}
  \bibinfo{person}{Sunghun Kim}.} \bibinfo{year}{2024}\natexlab{}.
\newblock \bibinfo{title}{A Survey on Large Language Models for Code
  Generation}.
\newblock
\showeprint[arxiv]{2406.00515}~[cs.CL]
\urldef\tempurl%
\url{https://arxiv.org/abs/2406.00515}
\showURL{%
\tempurl}


\bibitem[Koza(1992)]%
        {koza:book}
\bibfield{author}{\bibinfo{person}{John~R. Koza}.}
  \bibinfo{year}{1992}\natexlab{}.
\newblock \bibinfo{booktitle}{\emph{Genetic Programming: On the Programming of
  Computers by Means of Natural Selection}}.
\newblock \bibinfo{publisher}{MIT Press}, \bibinfo{address}{Cambridge, MA,
  USA}.
\newblock
\showISBNx{0-262-11170-5}
\urldef\tempurl%
\url{http://mitpress.mit.edu/books/genetic-programming}
\showURL{%
\tempurl}


\bibitem[Kumar et~al\mbox{.}(2026)]%
        {digitalredqueen}
\bibfield{author}{\bibinfo{person}{Akarsh Kumar}, \bibinfo{person}{Ryan
  Bahlous-Boldi}, \bibinfo{person}{Prafull Sharma}, \bibinfo{person}{Phillip
  Isola}, \bibinfo{person}{Sebastian Risi}, \bibinfo{person}{Yujin Tang}, {and}
  \bibinfo{person}{David Ha}.} \bibinfo{year}{2026}\natexlab{}.
\newblock \bibinfo{title}{Digital Red Queen: Adversarial Program Evolution in
  Core War with LLMs}.
\newblock
\showeprint[arxiv]{2601.03335}~[cs.AI]
\urldef\tempurl%
\url{https://arxiv.org/abs/2601.03335}
\showURL{%
\tempurl}


\bibitem[Lange et~al\mbox{.}(2024)]%
        {LLM_ES}
\bibfield{author}{\bibinfo{person}{Robert Lange}, \bibinfo{person}{Yingtao
  Tian}, {and} \bibinfo{person}{Yujin Tang}.} \bibinfo{year}{2024}\natexlab{}.
\newblock \showarticletitle{Large Language Models As Evolution Strategies}. In
  \bibinfo{booktitle}{\emph{Proceedings of the Genetic and Evolutionary
  Computation Conference Companion}} (Melbourne, VIC, Australia)
  \emph{(\bibinfo{series}{GECCO '24 Companion})}.
  \bibinfo{publisher}{Association for Computing Machinery},
  \bibinfo{address}{New York, NY, USA}, \bibinfo{pages}{579–582}.
\newblock
\showISBNx{9798400704956}
\href{https://doi.org/10.1145/3638530.3654238}{doi:\nolinkurl{10.1145/3638530.3654238}}


\bibitem[Lange et~al\mbox{.}(2026)]%
        {shinkaevolve}
\bibfield{author}{\bibinfo{person}{Robert~Tjarko Lange}, \bibinfo{person}{Yuki
  Imajuku}, {and} \bibinfo{person}{Edoardo Cetin}.}
  \bibinfo{year}{2026}\natexlab{}.
\newblock \showarticletitle{ShinkaEvolve: Towards Open-Ended and
  Sample-Efficient Program Evolution}. In \bibinfo{booktitle}{\emph{The
  Fourteenth International Conference on Learning Representations}}.
\newblock
\urldef\tempurl%
\url{https://openreview.net/forum?id=lKEdGCoDNC}
\showURL{%
\tempurl}


\bibitem[Lehman et~al\mbox{.}(2024)]%
        {ELM}
\bibfield{author}{\bibinfo{person}{Joel Lehman}, \bibinfo{person}{Jonathan
  Gordon}, \bibinfo{person}{Shawn Jain}, \bibinfo{person}{Kamal Ndousse},
  \bibinfo{person}{Cathy Yeh}, {and} \bibinfo{person}{Kenneth~O. Stanley}.}
  \bibinfo{year}{2024}\natexlab{}.
\newblock \bibinfo{booktitle}{\emph{Evolution Through Large Models}}.
\newblock \bibinfo{publisher}{Springer Nature Singapore},
  \bibinfo{address}{Singapore}, \bibinfo{pages}{331--366}.
\newblock
\showISBNx{978-981-99-3814-8}
\href{https://doi.org/10.1007/978-981-99-3814-8_11}{doi:\nolinkurl{10.1007/978-981-99-3814-8_11}}


\bibitem[Liu et~al\mbox{.}(2024)]%
        {LLM_EOH}
\bibfield{author}{\bibinfo{person}{Fei Liu}, \bibinfo{person}{Xialiang Tong},
  \bibinfo{person}{Mingxuan Yuan}, \bibinfo{person}{Xi Lin},
  \bibinfo{person}{Fu Luo}, \bibinfo{person}{Zhenkun Wang},
  \bibinfo{person}{Zhichao Lu}, {and} \bibinfo{person}{Qingfu Zhang}.}
  \bibinfo{year}{2024}\natexlab{}.
\newblock \showarticletitle{Evolution of heuristics: towards efficient
  automatic algorithm design using large language model}. In
  \bibinfo{booktitle}{\emph{Proceedings of the 41st International Conference on
  Machine Learning}} (Vienna, Austria) \emph{(\bibinfo{series}{ICML'24})}.
  \bibinfo{publisher}{JMLR.org}, Article \bibinfo{articleno}{1304},
  \bibinfo{numpages}{23}~pages.
\newblock


\bibitem[Lu et~al\mbox{.}(2024a)]%
        {LLM_discopop}
\bibfield{author}{\bibinfo{person}{Chris Lu}, \bibinfo{person}{Samuel Holt},
  \bibinfo{person}{Claudio Fanconi}, \bibinfo{person}{Alex~J. Chan},
  \bibinfo{person}{Jakob Foerster}, \bibinfo{person}{Mihaela van~der Schaar},
  {and} \bibinfo{person}{Robert~Tjarko Lange}.}
  \bibinfo{year}{2024}\natexlab{a}.
\newblock \showarticletitle{Discovering Preference Optimization Algorithms with
  and for Large Language Models}. In \bibinfo{booktitle}{\emph{Advances in
  Neural Information Processing Systems}},
  \bibfield{editor}{\bibinfo{person}{A.~Globerson},
  \bibinfo{person}{L.~Mackey}, \bibinfo{person}{D.~Belgrave},
  \bibinfo{person}{A.~Fan}, \bibinfo{person}{U.~Paquet},
  \bibinfo{person}{J.~Tomczak}, {and} \bibinfo{person}{C.~Zhang}} (Eds.),
  Vol.~\bibinfo{volume}{37}. \bibinfo{publisher}{Curran Associates, Inc.},
  \bibinfo{pages}{86528--86573}.
\newblock
\urldef\tempurl%
\url{https://proceedings.neurips.cc/paper_files/paper/2024/file/9d88b87b31986f8293bb0067a841579e-Paper-Conference.pdf}
\showURL{%
\tempurl}


\bibitem[Lu et~al\mbox{.}(2024b)]%
        {LLM_aiscientist}
\bibfield{author}{\bibinfo{person}{Chris Lu}, \bibinfo{person}{Cong Lu},
  \bibinfo{person}{Robert~Tjarko Lange}, \bibinfo{person}{Jakob Foerster},
  \bibinfo{person}{Jeff Clune}, {and} \bibinfo{person}{David Ha}.}
  \bibinfo{year}{2024}\natexlab{b}.
\newblock \bibinfo{title}{The AI Scientist: Towards Fully Automated Open-Ended
  Scientific Discovery}.
\newblock
\showeprint[arxiv]{2408.06292}~[cs.AI]
\urldef\tempurl%
\url{https://arxiv.org/abs/2408.06292}
\showURL{%
\tempurl}


\bibitem[Ma et~al\mbox{.}(2024)]%
        {LLM_eureka}
\bibfield{author}{\bibinfo{person}{Yecheng~Jason Ma}, \bibinfo{person}{William
  Liang}, \bibinfo{person}{Guanzhi Wang}, \bibinfo{person}{De-An Huang},
  \bibinfo{person}{Osbert Bastani}, \bibinfo{person}{Dinesh Jayaraman},
  \bibinfo{person}{Yuke Zhu}, \bibinfo{person}{Linxi Fan}, {and}
  \bibinfo{person}{Anima Anandkumar}.} \bibinfo{year}{2024}\natexlab{}.
\newblock \showarticletitle{Eureka: Human-Level Reward Design via Coding Large
  Language Models}. In \bibinfo{booktitle}{\emph{The Twelfth International
  Conference on Learning Representations}}.
\newblock
\urldef\tempurl%
\url{https://openreview.net/forum?id=IEduRUO55F}
\showURL{%
\tempurl}


\bibitem[Madaan et~al\mbox{.}(2023)]%
        {llm_selfrefine23}
\bibfield{author}{\bibinfo{person}{Aman Madaan}, \bibinfo{person}{Niket
  Tandon}, \bibinfo{person}{Prakhar Gupta}, \bibinfo{person}{Skyler Hallinan},
  \bibinfo{person}{Luyu Gao}, \bibinfo{person}{Sarah Wiegreffe},
  \bibinfo{person}{Uri Alon}, \bibinfo{person}{Nouha Dziri},
  \bibinfo{person}{Shrimai Prabhumoye}, \bibinfo{person}{Yiming Yang},
  \bibinfo{person}{Shashank Gupta}, \bibinfo{person}{Bodhisattwa~Prasad
  Majumder}, \bibinfo{person}{Katherine Hermann}, \bibinfo{person}{Sean
  Welleck}, \bibinfo{person}{Amir Yazdanbakhsh}, {and} \bibinfo{person}{Peter
  Clark}.} \bibinfo{year}{2023}\natexlab{}.
\newblock \showarticletitle{Self-Refine: Iterative Refinement with
  Self-Feedback}. In \bibinfo{booktitle}{\emph{Thirty-seventh Conference on
  Neural Information Processing Systems}}.
\newblock
\urldef\tempurl%
\url{https://openreview.net/forum?id=S37hOerQLB}
\showURL{%
\tempurl}


\bibitem[Meyerson et~al\mbox{.}(2024)]%
        {LMX}
\bibfield{author}{\bibinfo{person}{Elliot Meyerson}, \bibinfo{person}{Mark~J.
  Nelson}, \bibinfo{person}{Herbie Bradley}, \bibinfo{person}{Adam Gaier},
  \bibinfo{person}{Arash Moradi}, \bibinfo{person}{Amy~K. Hoover}, {and}
  \bibinfo{person}{Joel Lehman}.} \bibinfo{year}{2024}\natexlab{}.
\newblock \showarticletitle{Language Model Crossover: Variation through
  Few-Shot Prompting}.
\newblock \bibinfo{journal}{\emph{ACM Trans. Evol. Learn. Optim.}}
  \bibinfo{volume}{4}, \bibinfo{number}{4}, Article \bibinfo{articleno}{27}
  (\bibinfo{date}{Nov.} \bibinfo{year}{2024}), \bibinfo{numpages}{40}~pages.
\newblock
\href{https://doi.org/10.1145/3694791}{doi:\nolinkurl{10.1145/3694791}}


\bibitem[Mohamed et~al\mbox{.}(2025)]%
        {mohamed2025brokentelephone}
\bibfield{author}{\bibinfo{person}{Amr Mohamed}, \bibinfo{person}{Mingmeng
  Geng}, \bibinfo{person}{Michalis Vazirgiannis}, {and} \bibinfo{person}{Guokan
  Shang}.} \bibinfo{year}{2025}\natexlab{}.
\newblock \showarticletitle{{LLM} as a Broken Telephone: Iterative Generation
  Distorts Information}. In \bibinfo{booktitle}{\emph{Proceedings of the 63rd
  Annual Meeting of the Association for Computational Linguistics (Volume 1:
  Long Papers)}}, \bibfield{editor}{\bibinfo{person}{Wanxiang Che},
  \bibinfo{person}{Joyce Nabende}, \bibinfo{person}{Ekaterina Shutova}, {and}
  \bibinfo{person}{Mohammad~Taher Pilehvar}} (Eds.).
  \bibinfo{publisher}{Association for Computational Linguistics},
  \bibinfo{address}{Vienna, Austria}, \bibinfo{pages}{7493--7509}.
\newblock
\showISBNx{979-8-89176-251-0}
\href{https://doi.org/10.18653/v1/2025.acl-long.371}{doi:\nolinkurl{10.18653/v1/2025.acl-long.371}}


\bibitem[Mollah et~al\mbox{.}(2025)]%
        {mollah2025visualtelephone}
\bibfield{author}{\bibinfo{person}{Sabbir Mollah}, \bibinfo{person}{Rohit
  Gupta}, \bibinfo{person}{Sirnam Swetha}, \bibinfo{person}{Qingyang Liu},
  \bibinfo{person}{Ahnaf Munir}, {and} \bibinfo{person}{Mubarak Shah}.}
  \bibinfo{year}{2025}\natexlab{}.
\newblock \bibinfo{title}{The Telephone Game: Evaluating Semantic Drift in
  Unified Models}.
\newblock
\showeprint[arxiv]{2509.04438}~[cs.CV]
\urldef\tempurl%
\url{https://arxiv.org/abs/2509.04438}
\showURL{%
\tempurl}


\bibitem[Novikov et~al\mbox{.}(2025)]%
        {alphaevolve}
\bibfield{author}{\bibinfo{person}{Alexander Novikov},
  \bibinfo{person}{Ng{\^{a}}n Vu}, \bibinfo{person}{Marvin Eisenberger},
  \bibinfo{person}{Emilien Dupont}, \bibinfo{person}{Po{-}Sen Huang},
  \bibinfo{person}{Adam~Zsolt Wagner}, \bibinfo{person}{Sergey Shirobokov},
  \bibinfo{person}{Borislav Kozlovskii}, \bibinfo{person}{Francisco J.~R.
  Ruiz}, \bibinfo{person}{Abbas Mehrabian}, \bibinfo{person}{M.~Pawan Kumar},
  \bibinfo{person}{Abigail See}, \bibinfo{person}{Swarat Chaudhuri},
  \bibinfo{person}{George Holland}, \bibinfo{person}{Alex Davies},
  \bibinfo{person}{Sebastian Nowozin}, \bibinfo{person}{Pushmeet Kohli}, {and}
  \bibinfo{person}{Matej Balog}.} \bibinfo{year}{2025}\natexlab{}.
\newblock \showarticletitle{AlphaEvolve: {A} coding agent for scientific and
  algorithmic discovery}.
\newblock \bibinfo{journal}{\emph{CoRR}}  \bibinfo{volume}{abs/2506.13131}
  (\bibinfo{year}{2025}).
\newblock
\showeprint[arXiv]{2506.13131}
\href{https://doi.org/10.48550/ARXIV.2506.13131}{doi:\nolinkurl{10.48550/ARXIV.2506.13131}}


\bibitem[O'Reilly(1997)]%
        {OReilly1997GP-neutrality}
\bibfield{author}{\bibinfo{person}{U.-M. O'Reilly}.}
  \bibinfo{year}{1997}\natexlab{}.
\newblock \showarticletitle{Using a distance metric on genetic programs to
  understand genetic operators}. In \bibinfo{booktitle}{\emph{1997 IEEE
  International Conference on Systems, Man, and Cybernetics. Computational
  Cybernetics and Simulation}}, Vol.~\bibinfo{volume}{5}.
  \bibinfo{pages}{4092--4097 vol.5}.
\newblock
\href{https://doi.org/10.1109/ICSMC.1997.637337}{doi:\nolinkurl{10.1109/ICSMC.1997.637337}}


\bibitem[Peitek et~al\mbox{.}(2026)]%
        {peitek2026refactoring}
\bibfield{author}{\bibinfo{person}{Norman Peitek}, \bibinfo{person}{Julia
  Hess}, {and} \bibinfo{person}{Sven Apel}.} \bibinfo{year}{2026}\natexlab{}.
\newblock \bibinfo{title}{From Restructuring to Stabilization: A Large-Scale
  Experiment on Iterative Code Readability Refactoring with Large Language
  Models}.
\newblock
\showeprint[arxiv]{2602.21833}~[cs.SE]
\urldef\tempurl%
\url{https://arxiv.org/abs/2602.21833}
\showURL{%
\tempurl}


\bibitem[Perez et~al\mbox{.}(2024)]%
        {perez2024telephone}
\bibfield{author}{\bibinfo{person}{Jeremy Perez}, \bibinfo{person}{Corentin
  Leger}, \bibinfo{person}{Marcela Ovando-Tellez}, \bibinfo{person}{Chris
  Foulon}, \bibinfo{person}{Joan Dussauld}, \bibinfo{person}{Pierre-Yves
  Oudeyer}, {and} \bibinfo{person}{Clement Moulin-Frier}.}
  \bibinfo{year}{2024}\natexlab{}.
\newblock \bibinfo{title}{Cultural evolution in populations of Large Language
  Models}.
\newblock
\showeprint[arxiv]{2403.08882}~[cs.MA]
\urldef\tempurl%
\url{https://arxiv.org/abs/2403.08882}
\showURL{%
\tempurl}


\bibitem[Poli et~al\mbox{.}(2008)]%
        {GP_field_guide}
\bibfield{author}{\bibinfo{person}{Riccardo Poli}, \bibinfo{person}{William~B.
  Langdon}, {and} \bibinfo{person}{Nicholas~Freitag McPhee}.}
  \bibinfo{year}{2008}\natexlab{}.
\newblock \bibinfo{booktitle}{\emph{A Field Guide to Genetic Programming}}.
\newblock \bibinfo{publisher}{Lulu Press}.
\newblock
\showISBNx{978-1-4092-0073-4}
\urldef\tempurl%
\url{http://www.gp-field-guide.org.uk}
\showURL{%
\tempurl}
\newblock
\shownote{With contributions by John R. Koza}.


\bibitem[Romera-Paredes et~al\mbox{.}(2024)]%
        {LLM_funsearch}
\bibfield{author}{\bibinfo{person}{Bernardino Romera-Paredes},
  \bibinfo{person}{Mohammadamin Barekatain}, \bibinfo{person}{Alexander
  Novikov}, \bibinfo{person}{Matej Balog}, \bibinfo{person}{M.~Pawan Kumar},
  \bibinfo{person}{Emilien Dupont}, \bibinfo{person}{Francisco J.~R. Ruiz},
  \bibinfo{person}{Jordan~S. Ellenberg}, \bibinfo{person}{Pengming Wang},
  \bibinfo{person}{Omar Fawzi}, \bibinfo{person}{Pushmeet Kohli}, {and}
  \bibinfo{person}{Alhussein Fawzi}.} \bibinfo{year}{2024}\natexlab{}.
\newblock \showarticletitle{Mathematical discoveries from program search with
  large language models}.
\newblock \bibinfo{journal}{\emph{Nature}} \bibinfo{volume}{625},
  \bibinfo{number}{7995} (\bibinfo{date}{01 Jan} \bibinfo{year}{2024}),
  \bibinfo{pages}{468--475}.
\newblock
\showISSN{1476-4687}
\href{https://doi.org/10.1038/s41586-023-06924-6}{doi:\nolinkurl{10.1038/s41586-023-06924-6}}


\bibitem[Sadasivan et~al\mbox{.}(2024)]%
        {sadasivan2023recursive}
\bibfield{author}{\bibinfo{person}{Vinu~Sankar Sadasivan},
  \bibinfo{person}{Aounon Kumar}, \bibinfo{person}{Sriram Balasubramanian},
  \bibinfo{person}{Wenxiao Wang}, {and} \bibinfo{person}{Soheil Feizi}.}
  \bibinfo{year}{2024}\natexlab{}.
\newblock \bibinfo{title}{Can {AI}-Generated Text be Reliably Detected?}
\newblock
\urldef\tempurl%
\url{https://openreview.net/forum?id=NvSwR4IvLO}
\showURL{%
\tempurl}


\bibitem[Shumailov et~al\mbox{.}(2024a)]%
        {shumailov2023curse}
\bibfield{author}{\bibinfo{person}{Ilia Shumailov}, \bibinfo{person}{Zakhar
  Shumaylov}, \bibinfo{person}{Yiren Zhao}, \bibinfo{person}{Yarin Gal},
  \bibinfo{person}{Nicolas Papernot}, {and} \bibinfo{person}{Ross Anderson}.}
  \bibinfo{year}{2024}\natexlab{a}.
\newblock \bibinfo{title}{The Curse of Recursion: Training on Generated Data
  Makes Models Forget}.
\newblock
\showeprint[arxiv]{2305.17493}~[cs.LG]
\urldef\tempurl%
\url{https://arxiv.org/abs/2305.17493}
\showURL{%
\tempurl}


\bibitem[Shumailov et~al\mbox{.}(2024b)]%
        {shumailov2024nature}
\bibfield{author}{\bibinfo{person}{Ilia Shumailov}, \bibinfo{person}{Zakhar
  Shumaylov}, \bibinfo{person}{Yiren Zhao}, \bibinfo{person}{Nicolas Papernot},
  \bibinfo{person}{Ross Anderson}, {and} \bibinfo{person}{Yarin Gal}.}
  \bibinfo{year}{2024}\natexlab{b}.
\newblock \showarticletitle{AI models collapse when trained on recursively
  generated data}.
\newblock \bibinfo{journal}{\emph{Nature}} \bibinfo{volume}{631},
  \bibinfo{number}{8022} (\bibinfo{date}{01 Jul} \bibinfo{year}{2024}),
  \bibinfo{pages}{755--759}.
\newblock
\showISSN{1476-4687}
\href{https://doi.org/10.1038/s41586-024-07566-y}{doi:\nolinkurl{10.1038/s41586-024-07566-y}}


\bibitem[Song et~al\mbox{.}(2024)]%
        {LLM_BBO}
\bibfield{author}{\bibinfo{person}{Xingyou Song}, \bibinfo{person}{Yingtao
  Tian}, \bibinfo{person}{Robert~Tjarko Lange}, \bibinfo{person}{Chansoo Lee},
  \bibinfo{person}{Yujin Tang}, {and} \bibinfo{person}{Yutian Chen}.}
  \bibinfo{year}{2024}\natexlab{}.
\newblock \showarticletitle{Position: leverage foundational models for
  black-box optimization}. In \bibinfo{booktitle}{\emph{Proceedings of the 41st
  International Conference on Machine Learning}} (Vienna, Austria)
  \emph{(\bibinfo{series}{ICML'24})}. \bibinfo{publisher}{JMLR.org}, Article
  \bibinfo{articleno}{1878}, \bibinfo{numpages}{13}~pages.
\newblock


\bibitem[Stein and Bäck(2025)]%
        {llamea}
\bibfield{author}{\bibinfo{person}{Niki~van Stein} {and}
  \bibinfo{person}{Thomas Bäck}.} \bibinfo{year}{2025}\natexlab{}.
\newblock \showarticletitle{LLaMEA: A Large Language Model Evolutionary
  Algorithm for Automatically Generating Metaheuristics}.
\newblock \bibinfo{journal}{\emph{IEEE Transactions on Evolutionary
  Computation}} \bibinfo{volume}{29}, \bibinfo{number}{2}
  (\bibinfo{year}{2025}), \bibinfo{pages}{331--345}.
\newblock
\href{https://doi.org/10.1109/TEVC.2024.3497793}{doi:\nolinkurl{10.1109/TEVC.2024.3497793}}


\bibitem[Tacheny(2026)]%
        {tacheny2025agentic}
\bibfield{author}{\bibinfo{person}{Nicolas Tacheny}.}
  \bibinfo{year}{2026}\natexlab{}.
\newblock \bibinfo{title}{Geometric Dynamics of Agentic Loops in Large Language
  Models}.
\newblock
\showeprint[arxiv]{2512.10350}~[cs.LG]
\urldef\tempurl%
\url{https://arxiv.org/abs/2512.10350}
\showURL{%
\tempurl}


\bibitem[Tripto et~al\mbox{.}(2024)]%
        {tripto2024theseus}
\bibfield{author}{\bibinfo{person}{Nafis~Irtiza Tripto},
  \bibinfo{person}{Saranya Venkatraman}, \bibinfo{person}{Dominik Macko},
  \bibinfo{person}{Robert Moro}, \bibinfo{person}{Ivan Srba},
  \bibinfo{person}{Adaku Uchendu}, \bibinfo{person}{Thai Le}, {and}
  \bibinfo{person}{Dongwon Lee}.} \bibinfo{year}{2024}\natexlab{}.
\newblock \showarticletitle{A Ship of Theseus: Curious Cases of Paraphrasing in
  {LLM}-Generated Texts}. In \bibinfo{booktitle}{\emph{Proceedings of the 62nd
  Annual Meeting of the Association for Computational Linguistics (Volume 1:
  Long Papers)}}, \bibfield{editor}{\bibinfo{person}{Lun-Wei Ku},
  \bibinfo{person}{Andre Martins}, {and} \bibinfo{person}{Vivek Srikumar}}
  (Eds.). \bibinfo{publisher}{Association for Computational Linguistics},
  \bibinfo{address}{Bangkok, Thailand}, \bibinfo{pages}{6608--6625}.
\newblock
\href{https://doi.org/10.18653/v1/2024.acl-long.357}{doi:\nolinkurl{10.18653/v1/2024.acl-long.357}}


\bibitem[van Stein et~al\mbox{.}(2025)]%
        {vanstein2025codeevolgraphs}
\bibfield{author}{\bibinfo{person}{Niki van Stein}, \bibinfo{person}{Anna
  V.~Kononova}, \bibinfo{person}{Lars Kotthoff}, {and} \bibinfo{person}{Thomas
  B\"{a}ck}.} \bibinfo{year}{2025}\natexlab{}.
\newblock \showarticletitle{Code Evolution Graphs: Understanding Large Language
  Model Driven Design of Algorithms}. In \bibinfo{booktitle}{\emph{Proceedings
  of the Genetic and Evolutionary Computation Conference}} (NH Malaga Hotel,
  Malaga, Spain) \emph{(\bibinfo{series}{GECCO '25})}.
  \bibinfo{publisher}{Association for Computing Machinery},
  \bibinfo{address}{New York, NY, USA}, \bibinfo{pages}{943–951}.
\newblock
\showISBNx{9798400714658}
\href{https://doi.org/10.1145/3712256.3726328}{doi:\nolinkurl{10.1145/3712256.3726328}}


\bibitem[van Stein et~al\mbox{.}(2026)]%
        {vanstein2025behaviourspace}
\bibfield{author}{\bibinfo{person}{Niki van Stein}, \bibinfo{person}{Haoran
  Yin}, \bibinfo{person}{Anna~V. Kononova}, \bibinfo{person}{Thomas B{\"a}ck},
  {and} \bibinfo{person}{Gabriela Ochoa}.} \bibinfo{year}{2026}\natexlab{}.
\newblock \showarticletitle{Behaviour Space Analysis of {LLM}-Driven
  Meta-Heuristic Discovery}. In \bibinfo{booktitle}{\emph{Computational
  Intelligence}}, \bibfield{editor}{\bibinfo{person}{Francesco Marcelloni},
  \bibinfo{person}{Kurosh Madani}, \bibinfo{person}{Niki van Stein}, {and}
  \bibinfo{person}{Joaquim Filipe}} (Eds.). \bibinfo{publisher}{Springer Nature
  Switzerland}, \bibinfo{address}{Cham}, \bibinfo{pages}{367--385}.
\newblock
\showISBNx{978-3-032-15635-8}


\bibitem[Wang et~al\mbox{.}(2024)]%
        {wang2024voyager}
\bibfield{author}{\bibinfo{person}{Guanzhi Wang}, \bibinfo{person}{Yuqi Xie},
  \bibinfo{person}{Yunfan Jiang}, \bibinfo{person}{Ajay Mandlekar},
  \bibinfo{person}{Chaowei Xiao}, \bibinfo{person}{Yuke Zhu},
  \bibinfo{person}{Linxi Fan}, {and} \bibinfo{person}{Anima Anandkumar}.}
  \bibinfo{year}{2024}\natexlab{}.
\newblock \showarticletitle{Voyager: An Open-Ended Embodied Agent with Large
  Language Models}.
\newblock \bibinfo{journal}{\emph{Transactions on Machine Learning Research}}
  (\bibinfo{year}{2024}).
\newblock
\showISSN{2835-8856}
\urldef\tempurl%
\url{https://openreview.net/forum?id=ehfRiF0R3a}
\showURL{%
\tempurl}


\bibitem[Wang et~al\mbox{.}(2025)]%
        {wang2025attractorcycles}
\bibfield{author}{\bibinfo{person}{Zhilin Wang}, \bibinfo{person}{Yafu Li},
  \bibinfo{person}{Jianhao Yan}, \bibinfo{person}{Yu Cheng}, {and}
  \bibinfo{person}{Yue Zhang}.} \bibinfo{year}{2025}\natexlab{}.
\newblock \showarticletitle{Unveiling Attractor Cycles in Large Language
  Models: A Dynamical Systems View of Successive Paraphrasing}. In
  \bibinfo{booktitle}{\emph{Proceedings of the 63rd Annual Meeting of the
  Association for Computational Linguistics (Volume 1: Long Papers)}},
  \bibfield{editor}{\bibinfo{person}{Wanxiang Che}, \bibinfo{person}{Joyce
  Nabende}, \bibinfo{person}{Ekaterina Shutova}, {and}
  \bibinfo{person}{Mohammad~Taher Pilehvar}} (Eds.).
  \bibinfo{publisher}{Association for Computational Linguistics},
  \bibinfo{address}{Vienna, Austria}, \bibinfo{pages}{12740--12755}.
\newblock
\showISBNx{979-8-89176-251-0}
\href{https://doi.org/10.18653/v1/2025.acl-long.624}{doi:\nolinkurl{10.18653/v1/2025.acl-long.624}}


\bibitem[Wei et~al\mbox{.}(2022)]%
        {llm_chain_of_thought22}
\bibfield{author}{\bibinfo{person}{Jason Wei}, \bibinfo{person}{Xuezhi Wang},
  \bibinfo{person}{Dale Schuurmans}, \bibinfo{person}{Maarten Bosma},
  \bibinfo{person}{Brian Ichter}, \bibinfo{person}{Fei Xia},
  \bibinfo{person}{Ed~H. Chi}, \bibinfo{person}{Quoc~V. Le}, {and}
  \bibinfo{person}{Denny Zhou}.} \bibinfo{year}{2022}\natexlab{}.
\newblock \showarticletitle{Chain-of-thought prompting elicits reasoning in
  large language models}. In \bibinfo{booktitle}{\emph{Proceedings of the 36th
  International Conference on Neural Information Processing Systems}} (New
  Orleans, LA, USA) \emph{(\bibinfo{series}{NIPS '22})}.
  \bibinfo{publisher}{Curran Associates Inc.}, \bibinfo{address}{Red Hook, NY,
  USA}, Article \bibinfo{articleno}{1800}, \bibinfo{numpages}{14}~pages.
\newblock
\showISBNx{9781713871088}


\bibitem[Wu et~al\mbox{.}(2024)]%
        {LLM_EA_review}
\bibfield{author}{\bibinfo{person}{Xingyu Wu}, \bibinfo{person}{Sheng-Hao Wu},
  \bibinfo{person}{Jibin Wu}, \bibinfo{person}{Liang Feng}, {and}
  \bibinfo{person}{Kay~Chen Tan}.} \bibinfo{year}{2024}\natexlab{}.
\newblock \showarticletitle{Evolutionary Computation in the Era of Large
  Language Model: Survey and Roadmap}.
\newblock \bibinfo{journal}{\emph{IEEE Transactions on Evolutionary
  Computation}} (\bibinfo{year}{2024}), \bibinfo{pages}{1--1}.
\newblock
\href{https://doi.org/10.1109/TEVC.2024.3506731}{doi:\nolinkurl{10.1109/TEVC.2024.3506731}}


\bibitem[Yao et~al\mbox{.}(2023)]%
        {yao2023react}
\bibfield{author}{\bibinfo{person}{Shunyu Yao}, \bibinfo{person}{Jeffrey Zhao},
  \bibinfo{person}{Dian Yu}, \bibinfo{person}{Nan Du}, \bibinfo{person}{Izhak
  Shafran}, \bibinfo{person}{Karthik~R Narasimhan}, {and} \bibinfo{person}{Yuan
  Cao}.} \bibinfo{year}{2023}\natexlab{}.
\newblock \showarticletitle{ReAct: Synergizing Reasoning and Acting in Language
  Models}. In \bibinfo{booktitle}{\emph{The Eleventh International Conference
  on Learning Representations}}.
\newblock
\urldef\tempurl%
\url{https://openreview.net/forum?id=WE_vluYUL-X}
\showURL{%
\tempurl}


\bibitem[Ye et~al\mbox{.}(2024)]%
        {ReEvo}
\bibfield{author}{\bibinfo{person}{Haoran Ye}, \bibinfo{person}{Jiarui Wang},
  \bibinfo{person}{Zhiguang Cao}, \bibinfo{person}{Federico Berto},
  \bibinfo{person}{Chuanbo Hua}, \bibinfo{person}{Haeyeon Kim},
  \bibinfo{person}{Jinkyoo Park}, {and} \bibinfo{person}{Guojie Song}.}
  \bibinfo{year}{2024}\natexlab{}.
\newblock \showarticletitle{ReEvo: Large Language Models as Hyper-Heuristics
  with Reflective Evolution}. In \bibinfo{booktitle}{\emph{Advances in Neural
  Information Processing Systems}},
  \bibfield{editor}{\bibinfo{person}{A.~Globerson},
  \bibinfo{person}{L.~Mackey}, \bibinfo{person}{D.~Belgrave},
  \bibinfo{person}{A.~Fan}, \bibinfo{person}{U.~Paquet},
  \bibinfo{person}{J.~Tomczak}, {and} \bibinfo{person}{C.~Zhang}} (Eds.),
  Vol.~\bibinfo{volume}{37}. \bibinfo{publisher}{Curran Associates, Inc.},
  \bibinfo{pages}{43571--43608}.
\newblock
\href{https://doi.org/10.52202/079017-1381}{doi:\nolinkurl{10.52202/079017-1381}}


\bibitem[Yu and Miller(2001)]%
        {miller2001-neutrality}
\bibfield{author}{\bibinfo{person}{Tina Yu} {and} \bibinfo{person}{Julian~F.
  Miller}.} \bibinfo{year}{2001}\natexlab{}.
\newblock \showarticletitle{Neutrality and the Evolvability of Boolean Function
  Landscape}. In \bibinfo{booktitle}{\emph{Genetic Programming, 4th European
  Conference, EuroGP 2001, Lake Como, Italy, April 18-20, 2001, Proceedings}}
  \emph{(\bibinfo{series}{Lecture Notes in Computer Science})},
  \bibfield{editor}{\bibinfo{person}{Julian~F. Miller}, \bibinfo{person}{Marco
  Tomassini}, \bibinfo{person}{Pier~Luca Lanzi}, \bibinfo{person}{Conor Ryan},
  \bibinfo{person}{Andrea Tettamanzi}, {and} \bibinfo{person}{William~B.
  Langdon}} (Eds.). \bibinfo{publisher}{Springer}, \bibinfo{pages}{204--217}.
\newblock
\href{https://doi.org/10.1007/3-540-45355-5\_16}{doi:\nolinkurl{10.1007/3-540-45355-5\_16}}


\end{thebibliography}

\appendix

\section{Prompt and Representation Details}

This section documents the program representations and prompt infrastructure used throughout the experiments.

\subsection{AST Representations}
\label{app:ast}

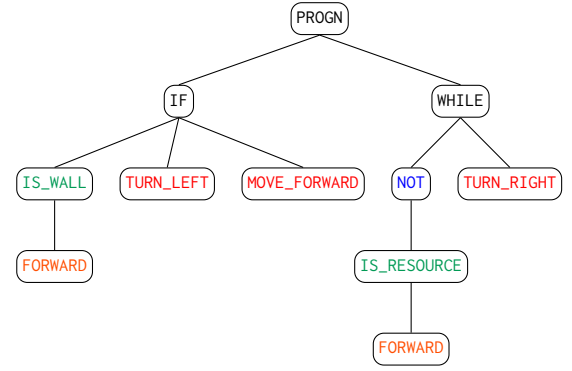
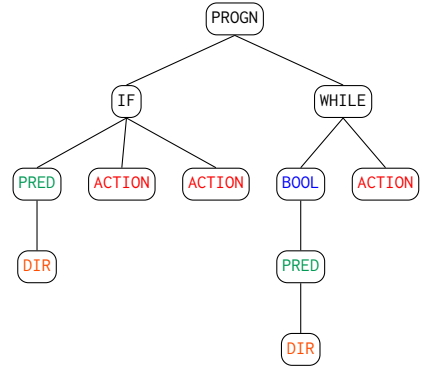
\begin{figure}
\centering
\begin{subfigure}{\columnwidth}
\centering
\begin{forest}
  for tree={
    draw,
    rounded corners,
    align=center,
    parent anchor=south,
    child anchor=north,
    l sep+=8pt,
    s sep+=4pt,
    inner sep=2pt,
    font=\ttfamily\footnotesize,
  }
[PROGN
  [IF
    [IS\_WALL, text=ForestGreen
      [FORWARD, text=red!70!yellow]
    ]
    [TURN\_LEFT, text=red]
    [MOVE\_FORWARD, text=red]
  ]
  [WHILE
    [NOT, text=blue
      [IS\_RESOURCE, text=ForestGreen
        [FORWARD, text=red!70!yellow]
      ]
    ]
    [TURN\_RIGHT, text=red]
  ]
]
\end{forest}
\caption{Program AST}
\label{fig:prog-ast}
\end{subfigure}

\vspace{0.3cm}
\begin{subfigure}{\columnwidth}
\centering
\begin{forest}
  for tree={
    draw,
    rounded corners,
    align=center,
    parent anchor=south,
    child anchor=north,
    l sep+=8pt,
    s sep+=4pt,
    inner sep=2pt,
    font=\ttfamily\footnotesize,
  }
[PROGN
  [IF
    [PRED, text=ForestGreen
      [DIR, text=red!70!yellow]
    ]
    [ACTION, text=red]
    [ACTION, text=red]
  ]
  [WHILE
    [BOOL, text=blue
      [PRED, text=ForestGreen
        [DIR, text=red!70!yellow]
      ]
    ]
    [ACTION, text=red]
  ]
]
\end{forest}
\caption{Skeleton AST}
\label{fig:skel-ast}
\end{subfigure}
\caption{AST representation of example program in the DSL (top) and its corresponding skeleton representation (bottom).}
\Description{Program Abstract Syntax Tree examples.}
\label{fig:dsl_tree}
\end{figure}

Figure~\ref{fig:dsl_tree} illustrates the abstract syntax tree (AST) representations used in the analysis. The full program AST preserves all terminal symbols, while the skeleton AST replaces terminals with their type categories (\texttt{ACTION}, \texttt{DIR}, \texttt{PRED}, \texttt{BOOL}), retaining only control-flow structure. This dual representation underlies the distinction between program-level and skeleton-level convergence reported in the main text.

\subsection{Mutation Prompt Templates}
\label{app:prompts}

Each mutation step sends the LLM a single prompt containing the DSL constraints, the current parent program, and an instruction line directing the model to produce a mutated variant.

The base mutation prompt template is:

\begin{spverbatim}
You are performing mutation on a genetic program for a grid-world agent.

Shared constraints:
<shared_info>

Parent program:
<parent_program>

Task:
- <variant_instruction>
- The produced program must be syntactically valid.
- Use only the allowed primitives and directions above.
- The program must include at least one action leaf (MOVE_FORWARD, TURN_LEFT, TURN_RIGHT).
- Do not include trailing commas.
- Output only the mutated program string.

\end{spverbatim}

\noindent The \textit{variant\_instruction} placeholder is the only component that varies across the 50 prompt conditions used in Experiment~1; the full list of instruction variants is provided in Appendix~\ref{app:prompt-generation}.

The \textit{shared\_info} block specifies the allowed DSL primitives and syntax rules. For the \texttt{no\_memory} primitive configuration used in all experiments, it renders as:

\begin{spverbatim}
Syntax: use uppercase function names with parentheses, e.g. IF(cond,then,else).
Each program must be a single expression. Output only the program string.

Allowed statements:
- IF(cond, then, else) - conditional branching.
- NO_OP - no operation (returns None).
- PROGN(stmt1, stmt2) - sequential evaluation.
- WHILE(cond, body) - loop with max-iteration guard.

Allowed action leaves:
- MOVE_FORWARD - action leaf.
- TURN_LEFT - action leaf.
- TURN_RIGHT - action leaf.

Allowed predicates:
- IS_EMPTY(DIR) - True if empty cell in DIR.
- IS_HAZARD(DIR) - True if a hazard in DIR.
- IS_RESOURCE(DIR) - True if a resource in DIR.
- IS_WALL(DIR) - True if a wall in DIR.

Allowed boolean ops:
- AND(a, b) - boolean conjunction.
- NOT(a) - boolean negation.
- OR(a, b) - boolean disjunction.

Allowed directions:
- FORWARD - relative direction.
- LEFT - relative direction.
- RIGHT - relative direction.

\end{spverbatim}

When a candidate program fails validation (e.g., uses a disallowed primitive or is syntactically malformed), the model is re-prompted with the following retry template:

\begin{spverbatim}
You are fixing a previously generated program that failed validation.

Use the allowed primitives and constraints below. Return only the corrected program string.

Failure reason:
<failure_reason>

Program to fix:
<candidate_text>

Constraints and allowed primitives:
<shared_info>

\end{spverbatim}

\subsection{Prompt Instruction Variants}
\label{app:prompt-generation}

The 50 mutation instruction variants used in Experiment~1 (Section~\ref{exp:prompt}) were generated by prompting ChatGPT to produce diverse paraphrasings of the core mutation command. Each variant replaces the \textit{variant\_instruction} placeholder in the base prompt template (Appendix~\ref{app:prompts}); all other prompt components remain identical across conditions. Table~\ref{tab:prompt-variants} lists all 50 variants. The four variants selected for Experiments~2 and~3 are indicated in bold, chosen based on their position in the distribution of unique program counts from Experiment~1: $\text{Prompt}_{\text{max}}$ produced the most unique programs, $\text{Prompt}_{\text{median}}$ fell at the median, $\text{Prompt}_{\text{prog-min}}$ produced the fewest unique programs, and $\text{Prompt}_{\text{skel-min}}$ produced the fewest unique skeletons. Below is the prompt that was used to produce the prompt variations.

\begin{spverbatim}
You will be generating 50 different variations of a specific instruction line that is used in genetic 
programming mutation operations. Each variation should convey the same core meaning but use different wording, phrasing, and terminology.

Here is the original instruction line you need to create variations of:

<original_line>
Produce a mutated program
</original_line>

Your task is to create 50 distinct variations of this instruction line. Follow these requirements carefully:

**Core Requirements:**
1. Each variation must convey the same fundamental meaning and instruction as the original line
2. Each variation must use different wording and phrasing from the original and from other variations
3. Each variation must maintain a clear, instructional tone appropriate for prompting an LLM
4. Each variation must be suitable for use in genetic programming mutation contexts
5. Each variation should be a complete, standalone instruction that could directly replace the original line

**Vocabulary and Terminology:**
- Vary the terminology you use across variations. Instead of always using "mutation," consider alternatives such as:
  - "change"
  - "variation" 
  - "modification"
  - "improvement"
  - "alteration"
  - "transformation"
  - "adjustment"
- Mix these terms across your variations to create lexical diversity

**Change Characterization:**
- Different variations should characterize the type or scope of changes differently. Consider including descriptors such as:
  - "small changes" or "minor modifications"
  - "creative changes" or "innovative variations"
  - "substantial improvements" or "significant alterations"
  - "incremental adjustments"
  - "exploratory modifications"
- Not every variation needs to specify the type of change, but several should to add useful diversity

**Output Format:**
After your planning, provide your 50 variations, each enclosed in numbered XML tags as follows:

<variation_1>
[First variation here]
</variation_1>

<variation_2>
[Second variation here]
</variation_2>

[Continue through variation_50]

\end{spverbatim}

\begin{table}[!htbp]
\caption{Labels of chosen representative prompts and corresponding IDs}
    \centering
    \begin{tabular}{ll}
    \toprule
        Prompt Label & Prompt ID \\
        \midrule
        $\text{Prompt}_{\text{max}}$      & 27 \\
        $\text{Prompt}_{\text{median}}$   & 43 \\
        $\text{Prompt}_{\text{prog-min}}$ & 47 \\
        $\text{Prompt}_{\text{skel-min}}$ & 21 \\
         \bottomrule
    \end{tabular}
    \label{tab:prompt-id}
\end{table}

\begin{table*}[!htbp]
\caption{All 50 mutation instruction variants. Variants selected for Experiments~2 and~3 are shown in bold with their paper labels.}
\centering
\small
\begin{tabular}{r p{0.42\textwidth} r p{0.42\textwidth}}
\toprule
ID & Instruction & ID & Instruction \\
\midrule
1 & Generate a modified version of the program. &
26 & Produce a new program instance with minor alterations. \\
2 & Create a new program that incorporates a mutation. &
\textbf{27} & \textbf{Generate a program that includes exploratory modifications.} \\
3 & Produce a slightly altered version of the given program. &
28 & Create a transformed variant of the program with adjustments. \\
4 & Construct a variation of the program with minor changes. &
29 & Produce a program that has undergone a mutation operation. \\
5 & Generate a transformed version of the program. &
30 & Generate a revised program incorporating small changes. \\
6 & Create a program that results from mutating the original code. &
31 & Create a modified program with slight structural differences. \\
7 & Produce an adjusted version of the program with incremental changes. &
32 & Produce a variation of the original program through alteration. \\
8 & Generate a creatively modified program derived from the original. &
33 & Generate a program that reflects a subtle transformation. \\
9 & Construct a revised version of the program with exploratory alterations. &
34 & Create a program with a controlled degree of modification. \\
10 & Create a new variant of the program through modification. &
35 & Produce a program variant obtained via mutation. \\
11 & Produce a slightly changed program based on the original. &
36 & Generate a modified program instance with incremental adjustments. \\
12 & Generate a program that reflects a mutation of the input program. &
37 & Create a program that introduces small changes to the original. \\
13 & Create an altered version of the program with small adjustments. &
38 & Produce a new program with slight deviations from the original. \\
14 & Produce a modified program that differs from the original. &
39 & Generate a program reflecting a creative alteration of the original. \\
15 & Generate a program variant that introduces a change to the existing structure. &
40 & Create a variation of the program with deliberate modifications. \\
16 & Create a transformed program by applying a mutation to the original code. &
41 & Produce a program that has been adjusted through mutation. \\
17 & Produce a revised program containing minor modifications. &
42 & Generate a program variant with refined changes. \\
18 & Generate an alternative version of the program with structural changes. &
\textbf{43} & \textbf{Create a program with minor but meaningful modifications.}  \\
19 & Create a program variation that modifies the original implementation. &
44 & Produce a transformed program reflecting a mutation step. \\
20 & Produce a new version of the program that includes an alteration. &
45 & Generate a program that includes a set of controlled alterations. \\
\textbf{21} & \textbf{Generate a program with a small mutation applied.}  &
46 & Create a modified version of the original program with small improvements. \\
22 & Create a slightly modified instance of the program. &
\textbf{47} & \textbf{Produce a program that has been slightly restructured.}  \\
23 & Produce a program that has been altered from its original form. &
48 & Generate a variant of the program through systematic modification. \\
24 & Generate a variation of the program with controlled changes. &
49 & Create a program that reflects a mutation-driven change. \\
25 & Create a program reflecting an incremental mutation step. &
50 & Produce a program incorporating an intentional alteration. \\
\bottomrule
\end{tabular}
\label{tab:prompt-variants}
\end{table*}

\section{Experimental Parameters}
\label{app:exps}

This section provides the initial programs and per-model configurations referenced in the experimental design.

\subsection{Initial Programs}

Three initial programs of varying complexity were generated using ramped half-and-half initialization with different random seeds. $C_{\text{medium}}$ serves as the standard initial program across all three experiments; Experiment~1 additionally uses $C_{\text{small}}$ and $C_{\text{large}}$ to assess sensitivity to initial program size.

\begin{figure}[!htbp]
  \centering
  \begin{subfigure}{0.9\columnwidth}
    \begin{lstlisting}[language=DSL, basicstyle=\ttfamily\small, breaklines=true]
MOVE_FORWARD
\end{lstlisting}
    \caption{$C_{\text{small}}$ (small initial program, seed 0).}
  \end{subfigure}
  \vspace{0.1cm}
   \begin{subfigure}{0.9\columnwidth}
    \begin{lstlisting}[language=DSL, basicstyle=\ttfamily\small, breaklines=true]
WHILE(IS_EMPTY(LEFT),
  PROGN(
    PROGN(PROGN(MOVE_FORWARD, TURN_LEFT), TURN_LEFT),
    IF(AND(IS_RESOURCE(LEFT), IS_WALL(RIGHT)),
       WHILE(IS_EMPTY(FORWARD), TURN_LEFT),
       IF(IS_EMPTY(FORWARD), TURN_RIGHT, MOVE_FORWARD))))
\end{lstlisting}
    \caption{$C_{\text{medium}}$ (medium initial program, seed 5).}
  \end{subfigure}
  \vspace{0.1cm}
  \begin{subfigure}{0.9\columnwidth}
    \begin{lstlisting}[language=DSL, basicstyle=\ttfamily\small, breaklines=true]
PROGN(
  IF(OR(OR(IS_HAZARD(RIGHT), IS_WALL(LEFT)),
        AND(IS_EMPTY(RIGHT), IS_HAZARD(LEFT))),
     IF(OR(IS_HAZARD(RIGHT), IS_HAZARD(FORWARD)),
        WHILE(IS_HAZARD(LEFT), NO_OP),
        IF(IS_HAZARD(LEFT), MOVE_FORWARD, TURN_RIGHT)),
     PROGN(PROGN(TURN_RIGHT, TURN_LEFT),
           PROGN(TURN_LEFT, TURN_RIGHT))),
  IF(AND(OR(IS_RESOURCE(FORWARD), IS_EMPTY(FORWARD)),
         OR(IS_EMPTY(RIGHT), IS_WALL(RIGHT))),
     PROGN(PROGN(NO_OP, TURN_LEFT), TURN_LEFT),
     PROGN(IF(IS_RESOURCE(RIGHT), MOVE_FORWARD, TURN_RIGHT),
           IF(IS_WALL(RIGHT), MOVE_FORWARD, MOVE_FORWARD))))
\end{lstlisting}
    \caption{$C_{\text{large}}$ (large initial program, seed 10).}
  \end{subfigure}
  \caption{Initial programs used in the experiments, generated via ramped half-and-half initialization with varying random seeds.}
  \Description{Three code listings showing DSL programs of increasing complexity. The small program is a single action (MOVE\_FORWARD). The medium program contains nested IF and WHILE statements with predicates and actions. The large program has deeply nested control flow with multiple boolean operators, predicates, and actions.}
  \label{fig:initial-programs}
\end{figure}

\subsection{Model Configurations}

Table~\ref{tab:model-configs} lists the per-model parameters for Experiment~3 (Section~\ref{exp:model}). Each model was tested with a primary provider and a fallback provider invoked when the primary exhausts its retry budget on a given mutation step. All models share the parameters in Table~\ref{tab:params} unless noted otherwise.

\begin{table*}[!htbp]
\caption{Per-model experimental parameters for Experiment~3. All models use temperature 1.0. The ``reasoning'' variants enable provider-specific reasoning modes (thinking level medium for Gemini, reasoning effort medium for OpenAI). $^\dagger$GPT-5 Mini (non-reasoning) uses an elevated retry budget of 15 and its fallback uses temperature 0.5 with reasoning effort medium to be able to complete chains of 300 iterations without failing.}
\centering
\small
\begin{tabular}{lllrrl}
\toprule
Model & Provider & Model ID & Tokens & Retries & Fallback \\
\midrule
Gemini 3.1 Flash Lite & Google & \texttt{gemini-3.1-flash-lite-preview} & 1024 & 5 & Gemini 3.1 Flash \\
Gemini 3.1 Flash Lite (reasoning) & Google & \texttt{gemini-3.1-flash-lite-preview} & 8192 & 5 & Gemini 3.1 Flash \\
Claude Haiku 4.5 & Anthropic & \texttt{claude-haiku-4-5-20251001} & 1024 & 5 & Claude Sonnet 4 \\
Claude Sonnet 4 & Anthropic & \texttt{claude-sonnet-4-20250514} & 1024 & 5 & Claude Sonnet 4 \\
Claude Sonnet 4.5 & Anthropic & \texttt{claude-sonnet-4-5-20250929} & 1024 & 5 & Claude Sonnet 4.5 \\
GPT-5 Mini$^\dagger$ & OpenAI & \texttt{gpt-5-mini-2025-08-07} & 4096 & 15 & GPT-5.1 \\
GPT-5 Mini (reasoning) & OpenAI & \texttt{gpt-5-mini-2025-08-07} & 12000 & 5 & GPT-5.1 \\
\bottomrule
\end{tabular}
\label{tab:model-configs}
\end{table*}

\section{Additional Results}
\label{sec:additional-results}

This section presents supplementary analyses that support the findings in the main text.

\begin{figure}[!htbp]
  \centering
  \includegraphics[width=0.9\linewidth]{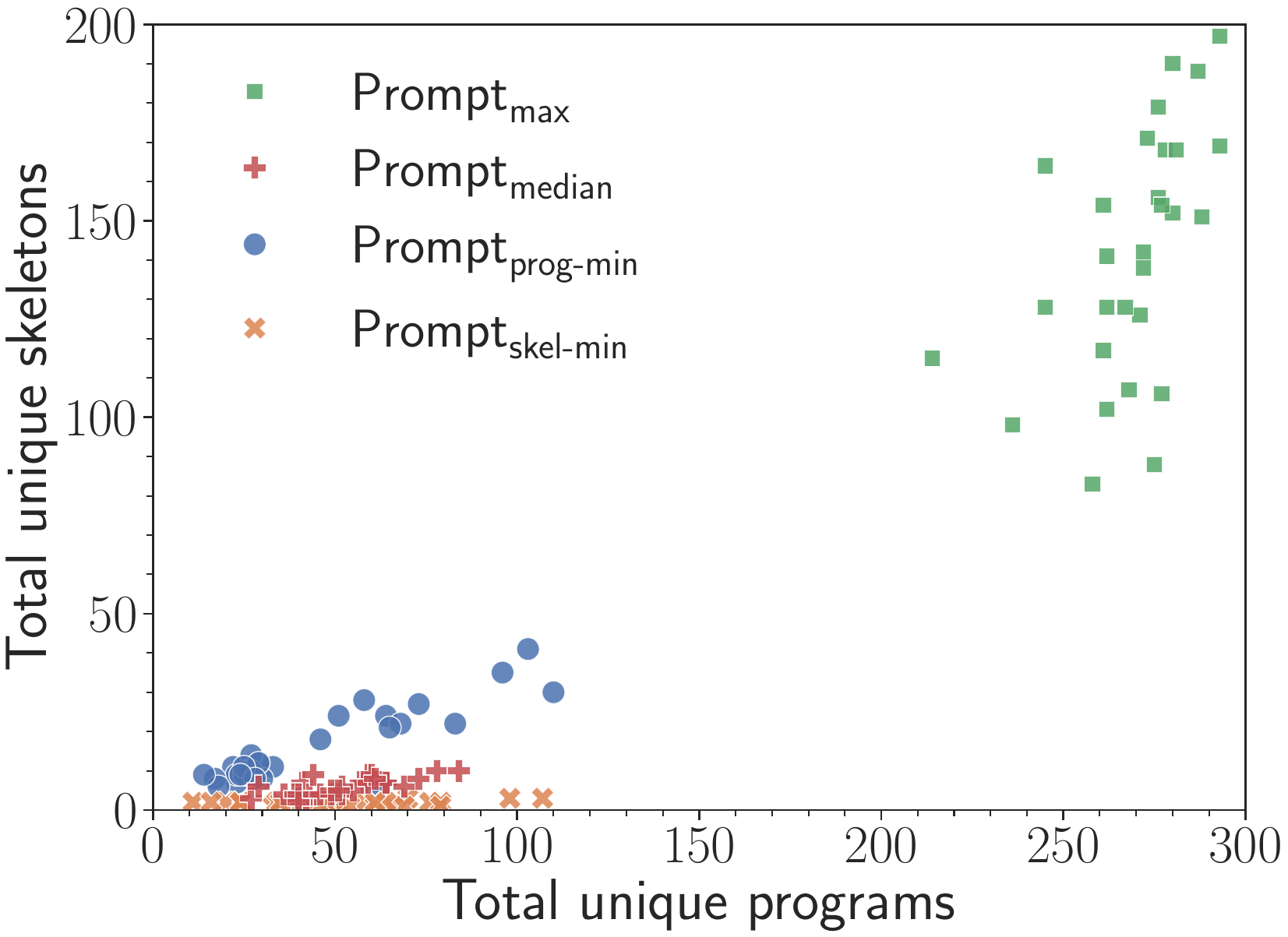}
  \caption{Total unique programs versus total unique skeletons for each of the 120 individual chains in Experiment 2, colored by prompt condition.}
  \Description{Scatter plot with unique programs on the x-axis and unique skeletons on the y-axis. Points cluster into distinct regions by prompt condition, with Prompt max in the upper-right and the remaining prompts compressed in the lower-left.}
  \label{fig:traj-scatter}
\end{figure}

Figure~\ref{fig:traj-scatter} plots total unique programs against total unique skeletons for each of the 120 individual chains. The endpoints of the trajectories stemming from these four prompts land in somewhat distinct clusters, with $\text{Prompt}_{\text{max}}$ occupying the upper-right region (high diversity on both axes) and the remaining prompts compressed into the lower-left. This separation reinforces the finding from Section~\ref{res:promptsweep}: prompt wording determines not just the quantity of exploration but its character, controlling whether variation extends to program structure or remains confined primarily to terminal substitutions.

\begin{figure}[!htbp]
    \centering
    \includegraphics[width=\linewidth]{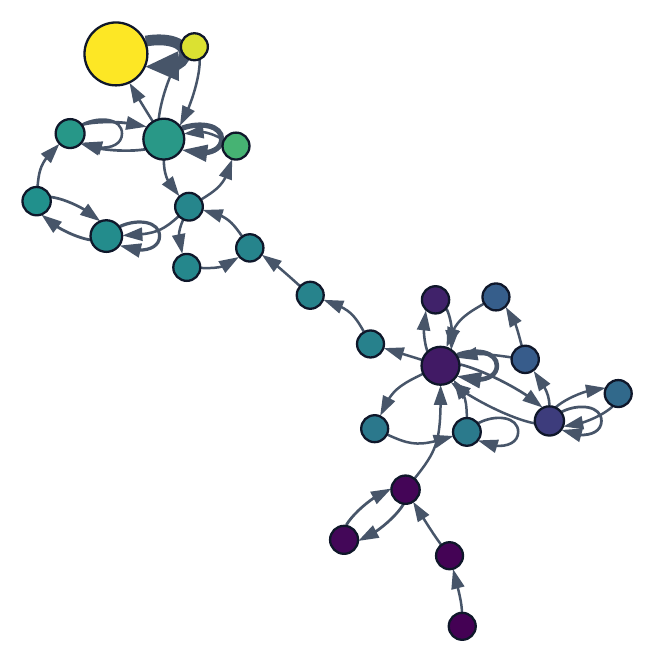}
    \caption{Skeleton-level transition graph corresponding to the same chain shown in Figure~\ref{fig:sample-mutation-trajectory} ($\text{Prompt}_{\text{median}}$, $C_{\text{medium}}$). At the skeleton level, the graph collapses to fewer nodes, with self-loops dominating the structure. Compare with the program-level graph in Figure~\ref{fig:sample-mutation-trajectory}, where the same chain produces a richer set of distinct states but still converges to a 2-cycle.}
    \Description{A directed graph showing skeleton states as nodes connected by edges representing mutations. The graph has fewer nodes than the program-level version, with prominent self-loops indicating that many mutations change only terminal values without altering the program's control-flow structure.}
    \label{fig:skeleton-transition-graph}
\end{figure}

\subsection{Prompt Sweep Full Results}
\label{app:prompt-sweep-results}

Table~\ref{tab:prompt-sweep-full} reports cumulative unique program and skeleton counts at step 300 for all 50 prompt variants and all three initial programs. Prompt variants selected for Experiments~2 and~3 are shown in bold.

Several entries in Table~\ref{tab:prompt-sweep-full} illustrate how prompts that read as semantically similar can produce markedly different convergence profiles. Five prompts that request small changes using ``slightly,'' ``minor,'' or ``small'' (variants 3, 4, 11, 13, and 17) each produce exactly 4 unique programs on $C_s$, while prompt 43 (``Create a program with minor but meaningful modifications'') produces 41; the qualifier ``but meaningful'' appears to shift behavior dramatically despite preserving the surface emphasis on small change. Similarly, prompts 15 (``introduces a change to the existing structure'') and 18 (``with structural changes'') both explicitly direct the operator toward structural variation, yet on $C_s$ produce 28 and 124 unique programs respectively.

\begin{table}[!htbp]
\caption{Cumulative unique programs and unique skeletons at step 300 for all 50 prompt variants across three initial programs. Prompt variants selected for Experiments~2 and~3 are shown in bold.}
\centering
\footnotesize
\begin{tabular}{r rr rr rr}
\toprule
& \multicolumn{2}{c}{$C_{\text{small}}$} & \multicolumn{2}{c}{$C_{\text{medium}}$} & \multicolumn{2}{c}{$C_{\text{large}}$} \\
\cmidrule(lr){2-3} \cmidrule(lr){4-5} \cmidrule(lr){6-7}
ID & Prog. & Skel. & Prog. & Skel. & Prog. & Skel. \\
\midrule
1  & 43  & 10  & 29  & 9  & 67  & 17  \\
2  & 11  & 6   & 72  & 8  & 197 & 10  \\
3  & 4   & 2   & 41  & 3  & 221 & 5   \\
4  & 4   & 2   & 24  & 5  & 155 & 5   \\
5  & 29  & 12  & 40  & 11 & 74  & 18  \\
6  & 51  & 8   & 43  & 5  & 164 & 12  \\
7  & 21  & 7   & 12  & 3  & 111 & 12  \\
8  & 190 & 75  & 178 & 59 & 218 & 85  \\
9  & 125 & 39  & 93  & 23 & 112 & 37  \\
10 & 38  & 13  & 30  & 9  & 43  & 17  \\
11 & 4   & 2   & 33  & 3  & 221 & 5   \\
12 & 60  & 9   & 41  & 4  & 226 & 12  \\
13 & 4   & 2   & 10  & 2  & 241 & 7   \\
14 & 14  & 5   & 13  & 2  & 146 & 6   \\
15 & 28  & 7   & 26  & 4  & 227 & 7   \\
16 & 25  & 7   & 26  & 3  & 199 & 10  \\
17 & 4   & 2   & 11  & 2  & 224 & 4   \\
18 & 124 & 47  & 135 & 55 & 133 & 48  \\
19 & 48  & 7   & 25  & 4  & 43  & 8   \\
20 & 22  & 6   & 47  & 5  & 154 & 5   \\
\textbf{21} & \textbf{37}  & \textbf{3}   & \textbf{33}  & \textbf{1}  & \textbf{224} & \textbf{3}   \\
22 & 6   & 3   & 41  & 2  & 182 & 5   \\
23 & 23  & 7   & 39  & 6  & 132 & 6   \\
24 & 46  & 7   & 38  & 5  & 78  & 8   \\
25 & 10  & 4   & 37  & 4  & 189 & 12  \\
26 & 6   & 3   & 62  & 2  & 225 & 6   \\
\textbf{27} & \textbf{249} & \textbf{142} & \textbf{162} & \textbf{86} & \textbf{278} & \textbf{168} \\
28 & 37  & 7   & 20  & 6  & 52  & 13  \\
29 & 55  & 8   & 45  & 3  & 158 & 20  \\
30 & 12  & 3   & 8   & 4  & 205 & 6   \\
31 & 47  & 15  & 26  & 4  & 73  & 27  \\
32 & 61  & 6   & 27  & 3  & 226 & 8   \\
33 & 8   & 4   & 71  & 4  & 82  & 5   \\
34 & 30  & 5   & 62  & 5  & 194 & 13  \\
35 & 84  & 10  & 24  & 4  & 220 & 9   \\
36 & 27  & 8   & 46  & 6  & 79  & 12  \\
37 & 22  & 6   & 64  & 5  & 227 & 4   \\
38 & 9   & 4   & 32  & 2  & 164 & 6   \\
39 & 134 & 43  & 120 & 49 & 140 & 47  \\
40 & 55  & 15  & 33  & 11 & 23  & 9   \\
41 & 61  & 8   & 97  & 14 & 138 & 9   \\
42 & 55  & 16  & 44  & 12 & 39  & 13  \\
\textbf{43} & \textbf{41}  & \textbf{10}  & \textbf{50}  & \textbf{7}  & \textbf{119} & \textbf{4}   \\
44 & 45  & 7   & 34  & 6  & 68  & 9   \\
45 & 92  & 21  & 74  & 9  & 197 & 8   \\
46 & 78  & 54  & 31  & 15 & 64  & 28  \\
\textbf{47} & \textbf{8}   & \textbf{5}   & \textbf{55}  & \textbf{19} & \textbf{28}  & \textbf{8}   \\
48 & 52  & 7   & 46  & 5  & 102 & 14  \\
49 & 80  & 16  & 58  & 11 & 59  & 10  \\
50 & 34  & 6   & 39  & 6  & 136 & 7   \\
\bottomrule
\end{tabular}
\label{tab:prompt-sweep-full}
\end{table}

\subsection{Pairwise Levenshtein Distance Heatmaps}
\label{sec:heatmaps}

\begin{figure}[!htbp]
    \centering
    \includegraphics[width=0.9\linewidth]{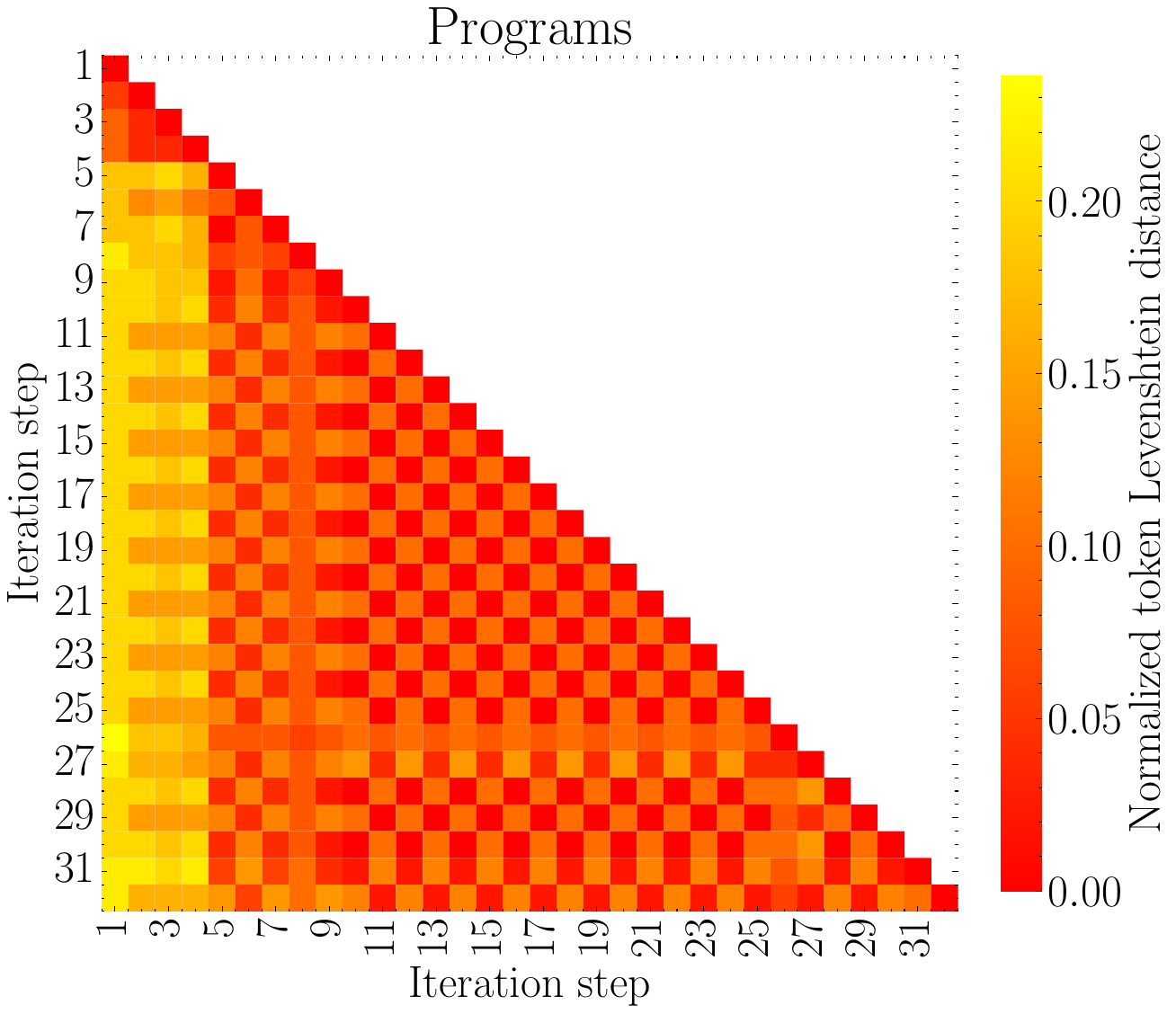}
    \caption{Pairwise normalized Levenshtein distance heatmap for a representative mutation chain, zoomed to the first 32 iterations. The checkered pattern along the diagonal indicates short-period cycling (primarily 2-cycles), while uniform dark blocks indicate intervals of stagnation where the chain revisited the same program repeatedly.}
    \label{fig:hm_zoom}
\end{figure}

\begin{figure}[!htbp]
    \centering
    \includegraphics[width=0.9\linewidth]{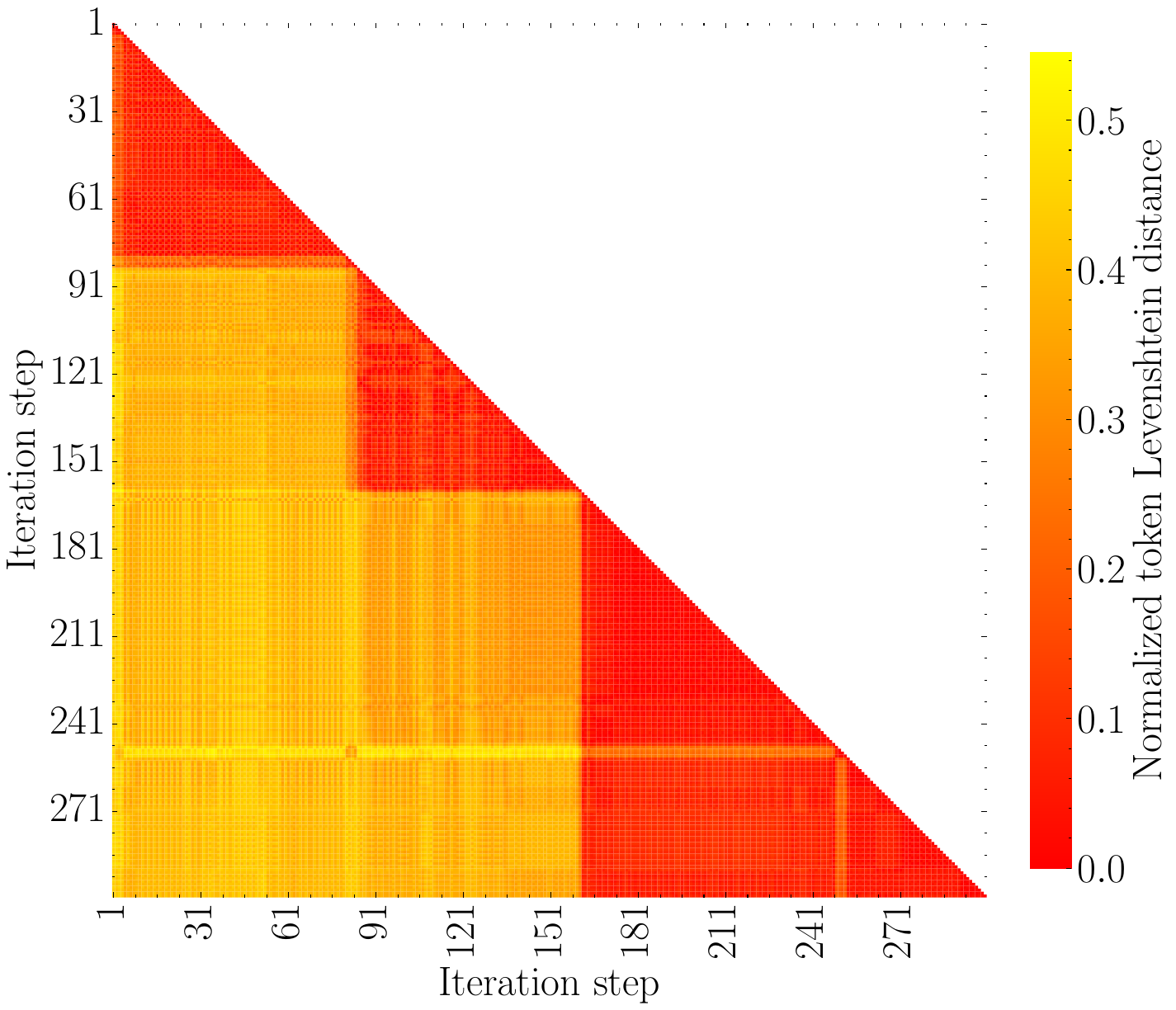}
    \caption{Full 300-iteration Levenshtein distance heatmap for the same chain as Figure~\ref{fig:hm_zoom}. Large dark blocks correspond to extended periods of stagnation, and the lighter off-diagonal rectangles between blocks mark transitions where the chain jumped to a structurally different program. The overall structure shows that convergence proceeds through discrete attractor-hopping rather than gradual contraction.}
    \label{fig:hm_full}
\end{figure}

To visualize the temporal structure of convergence at the individual-chain level, we compute pairwise normalized token-level Levenshtein distances between all programs at each pair of steps in a chain. Each heatmap is lower-triangular with iteration steps on both axes. Red regions indicate low distance between programs at those steps (revisitation of similar forms), while yellow regions indicate greater divergence.

\begin{figure*}[!htbp]
    \centering
    \includegraphics[width=0.65\linewidth]{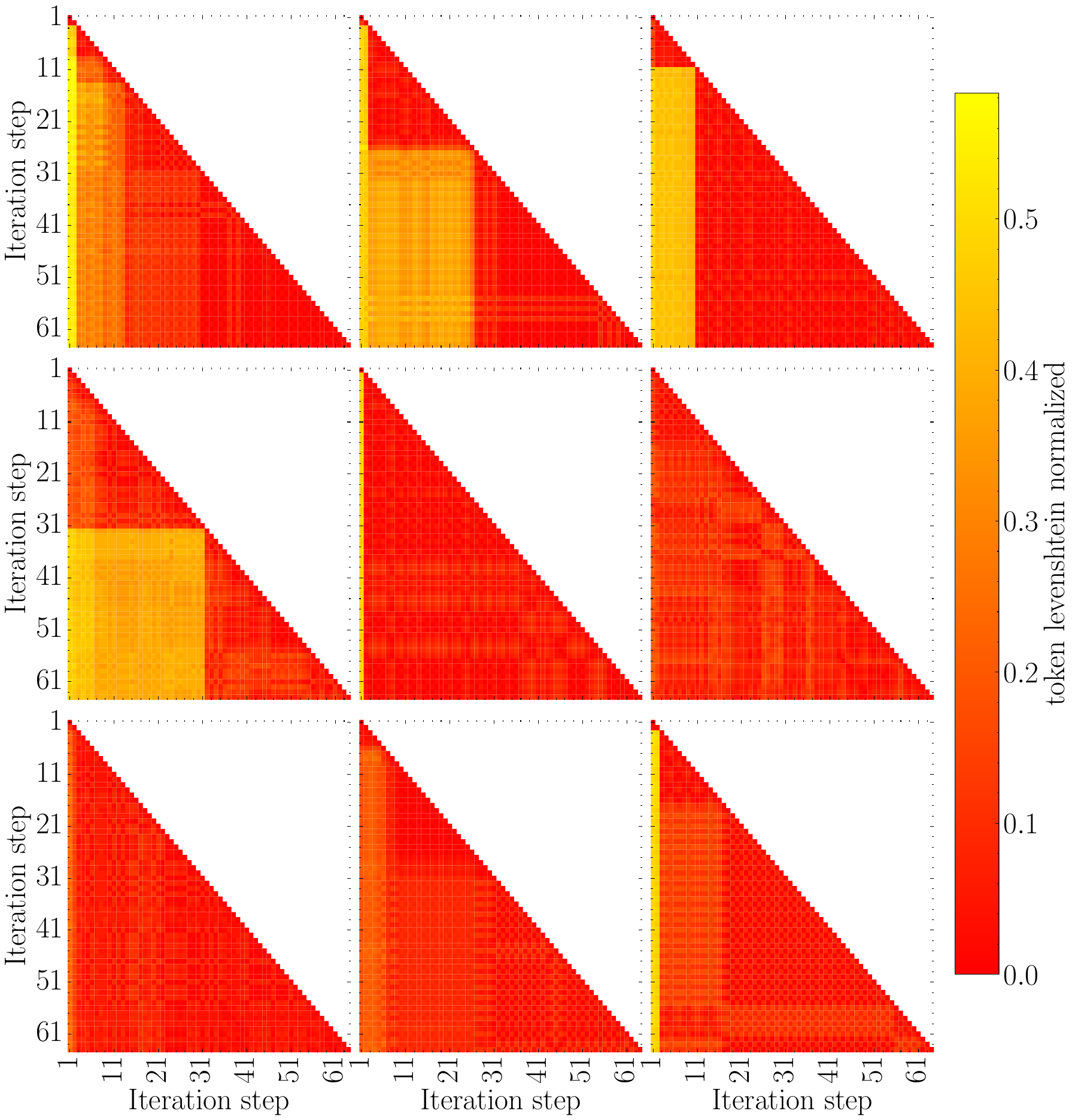}
    \caption{Levenshtein distance heatmaps for nine independent mutation chains (first 64 iterations each). Checkered patterns and block structures appear consistently across replications, though the specific block sizes and transition points vary. Some chains (e.g., top-left) enter stagnation almost immediately, while others (e.g., bottom-right) sustain more varied exploration before settling.}
    \label{fig:hm_grid}
\end{figure*}

Two recurring visual motifs are apparent across chains. The first is a \textit{checkered pattern}: alternating bands of low and high distance that produce a grid-like texture along the diagonal. This pattern arises when a chain oscillates between two (or a small number of) programs, so that even-numbered iterations are similar to one another and odd-numbered iterations form a separate cluster. The resulting structure is the spatial signature of short limit cycles, most commonly 2-cycles. Wang et al.~\cite{wang2025attractorcycles} document an analogous pattern in their difference confusion matrices for successive paraphrasing, where alternating light and dark bands reveal 2-period attractor cycles in natural-language trajectories. The checkered regions in our heatmaps provide direct visual evidence that the same periodic dynamics occur in LLM-driven mutation over program space, not only in textual paraphrasing.

The second motif consists of \textit{rectangular blocks} of uniform color. A dark (red) rectangle spanning iterations $i$ through $j$ on both axes indicates that the chain revisited the same program or a small set of nearly identical programs throughout that interval, a period of stagnation in which mutation produced no effective change. Conversely, a lighter (yellow) off-diagonal rectangle between two such blocks indicates a transition: the chain jumped from one attractor region to another, producing programs that differ substantially from those in the preceding block. Together, these block structures reveal that convergence is not always a smooth process. Chains can remain trapped in a narrow region of program space for dozens of iterations before abruptly shifting to a different region, only to stagnate again. The heatmaps thus complement the cumulative unique-count curves presented in Section~\ref{res:promptsweep} by exposing the temporal fine structure that aggregate statistics obscure.

Figure~\ref{fig:hm_zoom} shows a representative 32-iteration segment at high resolution, making the checkered and block structures clearly visible. Figure~\ref{fig:hm_full} shows the same chain over the full 300 iterations, where the large-scale block structure and transitions between attractor regions are prominent. Figure~\ref{fig:hm_grid} presents nine independent chains side by side, illustrating both the consistency of these patterns across replications and the variation in block size and transition frequency.

\end{document}